\documentclass{article} % For LaTeX2e
\usepackage{iclr2027_conference,times}
\usepackage{amsmath,amsfonts,bm}

\def\eqref#1{equation~\ref{#1}}
\def\1{\bm{1}}

\DeclareMathAlphabet{\mathsfit}{\encodingdefault}{\sfdefault}{m}{sl}
\SetMathAlphabet{\mathsfit}{bold}{\encodingdefault}{\sfdefault}{bx}{n}

\usepackage{hyperref}
\hypersetup{hidelinks}
\usepackage[table]{xcolor}
\definecolor{selfbg}{HTML}{EDF3FA}
\definecolor{secondbg}{HTML}{FCECEF}
\definecolor{bestbg}{HTML}{F8ECCD}
\definecolor{studentbg}{HTML}{EDF3FA}
\definecolor{teacherbg}{HTML}{CADCEF}
\definecolor{selfaccent}{HTML}{245A81} 
\definecolor{resultblue}{HTML}{245B78}
\usepackage[most]{tcolorbox}
\usepackage{xcolor}
\usepackage{amsmath}

\usepackage[most]{tcolorbox}
\usepackage{xcolor}

\definecolor{accountframe}{HTML}{A8A8A8}
\definecolor{accountbg}{HTML}{FAFAFA}

\newtcolorbox{accountbox}{
    enhanced,
    width=\linewidth,
    colback=accountbg,
    colframe=accountframe,
    boxrule=0.45pt,
    arc=1mm,
    outer arc=1mm,
    left=3.5mm,
    right=3.5mm,
    top=2.8mm,
    bottom=2.8mm,
    before skip=7pt,
    after skip=7pt
}

\definecolor{historybg}{HTML}{F2F2F2}
\definecolor{historylabel}{HTML}{303030}
\definecolor{historymeta}{HTML}{666666}

\newtcolorbox{historybox}{
    enhanced,
    breakable,
    width=\linewidth,
    colback=historybg,
    colframe=historybg,
    boxrule=0pt,
    arc=0mm,
    outer arc=0mm,
    left=5mm,
    right=5mm,
    top=4mm,
    bottom=4mm,
    fontupper=\ttfamily\small,
    before skip=6pt,
    after skip=6pt,
    pad at break*=2mm
}

\newcommand{\trajlabel}[1]{%
    \textbf{\textcolor{historylabel}{[#1]}}%
}

\definecolor{purplebg}{HTML}{E8E8FB}

\definecolor{greenbg}{HTML}{E8F8E8}

\definecolor{boxframe}{HTML}{303030}

\definecolor{metagray}{HTML}{555555}

\definecolor{purplebg}{HTML}{E8E8FB}
\definecolor{greenbg}{HTML}{E8F8E8}
\definecolor{boxframe}{HTML}{303030}
\definecolor{metagray}{HTML}{555555}

\newtcolorbox{predictionbox}[1]{
    enhanced,
    width=\linewidth,
    colback=#1,
    colframe=boxframe,
    boxrule=0.45pt,
    arc=0mm,
    outer arc=0mm,
    boxsep=0pt,
    left=3mm,
    right=3mm,
    top=2.5mm,
    bottom=2.5mm,
    fontupper=\ttfamily\small,
    before skip=0pt,
    after skip=0pt
}

\newcommand{\field}[1]{\textbf{#1}}
\newcommand{\status}[1]{\textbf{#1}}

\usepackage{calc}
\usepackage{url}
\usepackage{xcolor}

\usepackage{enumitem}
\usepackage{booktabs}
\usepackage{multirow}
\usepackage{algorithm}
\usepackage{algpseudocode}
\usepackage{pifont}
\usepackage{float}
\usepackage{graphicx}

\usepackage[export]{adjustbox}
\usepackage{sidecap}

\definecolor{propertygreen}{HTML}{356B28}
\definecolor{propertyred}{HTML}{A33238}

\usepackage[most]{tcolorbox}

\definecolor{promptred}{HTML}{AD0808}
\definecolor{promptgreen}{HTML}{079900}
\definecolor{promptblue}{HTML}{2463AD}

\title{
Environmental Feedback Modeling Matters: Rethinking Feedback Treatment in Agentic Hindsight Self-Distillation
}

\author{%
\textbf{Hangxi Guo}$^{1,*}$,
\textbf{Fengyuan Liu}$^{1,*}$,
\textbf{Yue Wang}$^{2,3}$,
\textbf{Yuhua Qi}$^{1}$,
\textbf{Haoyi Xiong}$^{4}$, \\
\textbf{Fei Sun}$^{5}$,
\textbf{Mengnan Du}$^{1,\dagger}$
\\
$^{1}$The Chinese University of Hong Kong, Shenzhen
$^{2}$Shanghai AI Laboratory
\\
$^{3}$University of Science and Technology of China
$^{4}$Independent
\\
$^{5}$Institute of Computing Technology, CAS \\
\texttt{keepstudying931@gmail.com, mengnandu@cuhk.edu.cn} \\
$^{*}$Equal contribution, $^{\dagger}$Corresponding author
}

\iclrfinalcopy

\begin{document}

\maketitle

\lhead{Preprint}

\begin{abstract}
Reinforcement learning is commonly
used to train language agents in interactive environments,
but cannot be directly applied when rewards are unavailable.
Recent methods use environmental feedback as privileged context
for hindsight self-distillation, but our analysis suggests that
simply conditioning the teacher on feedback is insufficient,
motivating us to rethink how environmental feedback is used in
agentic self-distillation.
Given that environmental feedback contains rich supervision
for modeling how the environment responds to agent actions,
we introduce \textit{agentic SElf-distilLation with
environmental Feedback modeling} (SELF), a framework that jointly
optimizes environmental feedback modeling and hindsight self-distillation.
SELF learns to predict environmental responses while
distilling guidance from a feedback-conditioned self-teacher
into the policy.
Our analysis reveals a mutually reinforcing mechanism:
environmental feedback modeling strengthens hindsight supervision
and policy learning, while self-distillation enhances the model's
ability to model environmental feedback.
With Qwen3-8B, SELF outperforms SDPO and GRPO by 6.4
and 4.1 percentage points in $\tau$-bench success rate,
and by 10.71 and 3.57 percentage points in AppWorld
task goal completion, respectively.
These results show that SELF uses environmental feedback
more effectively within agentic self-distillation,
improving agent capabilities.
\end{abstract}

\section{Introduction}

In recent years, training language agents to solve complex tasks
in interactive environments involving tool use and code execution
has become increasingly important~\citep{yao2023react,trivedi2024appworld,choudhury2025better}.
A common training approach is Reinforcement Learning with Verifiable
Rewards (RLVR)~\citep{lambert2025tulu3}, where algorithms such as Group Relative Policy
Optimization (GRPO)~\citep{shao2024deepseekmath} use trajectory-level rewards to optimize
the policy.
In some settings, however, reliable rewards are unavailable~\citep{tian2026autorubric,liu2026dynacf},
and environmental feedback becomes the primary source of
external information for agent training.
This feedback arises naturally from interactions and includes
tool outputs, code execution results, and environmental state
changes caused by the agent's actions.
Since such feedback typically describes action consequences
rather than providing direct supervision for policy learning,
using it effectively in agent training remains a challenge.

Recent methods tackle this challenge through hindsight
self-distillation.
SDPO~\citep{hubotter2026sdpo} conditions a self-teacher on
environmental feedback, while OEL~\citep{ye2026oel} further
extracts transferable knowledge from interaction experience.
Our analysis shows that, despite containing sufficient information
for policy improvement, such feedback yields only modest gains
when used in these ways, suggesting that the key limitation lies
in how it is exploited during training.
Although environmental feedback does not directly supervise
action generation, it provides direct targets for learning how
the environment responds to agent actions.
We therefore ask:
\textit{Can learning to predict environmental feedback enable
more effective use of it in agentic self-distillation?}

Our analysis reveals a mutually reinforcing relationship:
environmental feedback modeling improves hindsight
self-distillation, while self-distillation in turn strengthens
environmental feedback prediction.
Building on this finding, we introduce
\textit{agentic SElf-distilLation with environmental
Feedback} (\textbf{SELF}), a framework that jointly optimizes
environmental feedback modeling and hindsight self-distillation.
For each interaction transition, SELF uses the realized feedback
in two complementary ways: as a prediction target for learning
how the environment responds to the agent's action, and as
privileged context for a self-teacher whose action distribution
is distilled into the student policy.
By jointly learning from these two signals within a shared model,
SELF enables more effective use of environmental feedback
for policy improvement.

We evaluate SELF on $\tau$-bench Retail~\citep{yao2025taubench} for multi-turn
customer-service interactions with domain-specific tools,
and AppWorld~\citep{trivedi2024appworld} for code-based task completion across
applications.
Across Qwen3-4B-Instruct and Qwen3-8B (non-thinking)~\citep{yang2025qwen3}, SELF
outperforms feedback-learning baselines and reward-trained
GRPO~\citep{shao2024deepseekmath} in task success and goal completion.
With Qwen3-8B, SELF surpasses SDPO~\citep{hubotter2026sdpo} and GRPO by 6.4 and 4.1  percentage points
 in $\tau$-bench success rate, and by
10.71 and 3.57 percentage points in AppWorld task goal
completion, respectively, which demonstrates SELF's effectiveness
in turning environmental feedback into useful supervision
for policy improvement.
Further analyses show that SELF also retains strong
generalization capabilities beyond the training environments.

Our key contributions can be summarized as follows:
\begin{itemize}[leftmargin=10pt, topsep=-2pt, itemsep=1pt, partopsep=1pt, parsep=1pt]
    \item We identify a mutually reinforcing mechanism:
    environmental feedback modeling improves hindsight
    self-distillation, while self-distillation strengthens
    environmental feedback prediction.

    \item We propose SELF, an agent training framework
    that learns from environmental feedback by jointly
    optimizing environmental feedback modeling and hindsight
    self-distillation.

    \item We empirically show that SELF improves performance on
    $\tau$-bench Retail and AppWorld, outperforming both
    existing self-distillation approaches and reward-trained
    GRPO, which demonstrates its effectiveness in enabling
    agentic self-distillation to make better use of
    environmental feedback.
\end{itemize}

\section{Related Work}
\label{sec:related_work}

\textbf{Agentic Self-Distillation.}
Agentic self-distillation provides dense policy supervision
without requiring a separate stronger teacher.
OPSD~\citep{zhao2026opsd} constructs student and teacher
policies from the same model, conditioning the teacher on
privileged reference information.
SDPO~\citep{hubotter2026sdpo} instead conditions the
self-teacher on textual feedback and distills its
feedback-informed predictions into the policy.
SDFT~\citep{shenfeld2026sdft} uses demonstration-conditioned
self-distillation to acquire new knowledge and skills while
mitigating forgetting.
RLSD~\citep{yang2026rlsd} further combines self-distillation
with RLVR, using privileged teachers for token-level credit while retaining verifier rewards to determine the policy-update direction.

\textbf{Learning from Environmental Feedback.}
Environmental feedback contains rich information about the
realized consequences of agent actions, and recent methods
exploit this information in different forms.
OCSD~\citep{yang2026ocsd} contrasts replay contexts with and
without future observations to calibrate policy updates, while
HERO~\citep{liu2026hero} converts subsequent observations into
turn-level hindsight diagnoses.
OEL~\citep{ye2026oel} extracts transferable experiential
knowledge from interaction trajectories before distilling it
into the policy.
Kleine Buening et al.~\citep{buening2026aligning} use raw user
follow-up messages as hindsight information for self-distillation.
AHEAD~\citep{jin2026ahead} conditions the teacher on environment
feedback and supplements critical error steps with generated
corrective hints, whereas OpenClaw-RL~\citep{wang2026openclawrl}
converts next-state signals into process rewards or hindsight
guidance for online agent learning.
These approaches use realized feedback to construct or refine supervision for action generation.

\textbf{Environmental Feedback Modeling for Policy Learning.}
A complementary line of work directly learns the predictive
structure contained in interaction feedback.
ECHO~\citep{shrivastava2026echo} and
PaW~\citep{lu2026paw} incorporate environment prediction into
policy optimization, while AAWM~\citep{cai2026aawm} studies
decision-oriented modeling targets rather than directly
reconstructing future observations.
Early Experience~\citep{zhang2026earlyexperience} uses
next states produced by agent actions as supervision for
implicit world modeling and self-reflection.
Most closely related, RLTF~\citep{song2026rltf} studies both
feedback-conditioned self-distillation and feedback modeling,
where textual critiques are predicted as an auxiliary objective.
In contrast, SELF models environmental responses directly from interactive trajectories and uses the same realized feedback both as a prediction target and as privileged context for hindsight self-distillation. Our analysis further shows that these objectives mutually reinforce each other.

\section{Methodology}
\label{sec:method}

We present SELF, a framework that improves agentic self-distillation by explicitly modeling environmental feedback. We first formalize agent–environment interaction and the problem of learning from environmental feedback. We then describe on-policy hindsight self-distillation, which uses realized feedback as privileged information to construct policy supervision. Finally, we introduce the environmental feedback modeling objective and show how it is jointly optimized with hindsight self-distillation within the framework.

\begin{figure}[t]
    \centering
    \includegraphics[width=\linewidth]{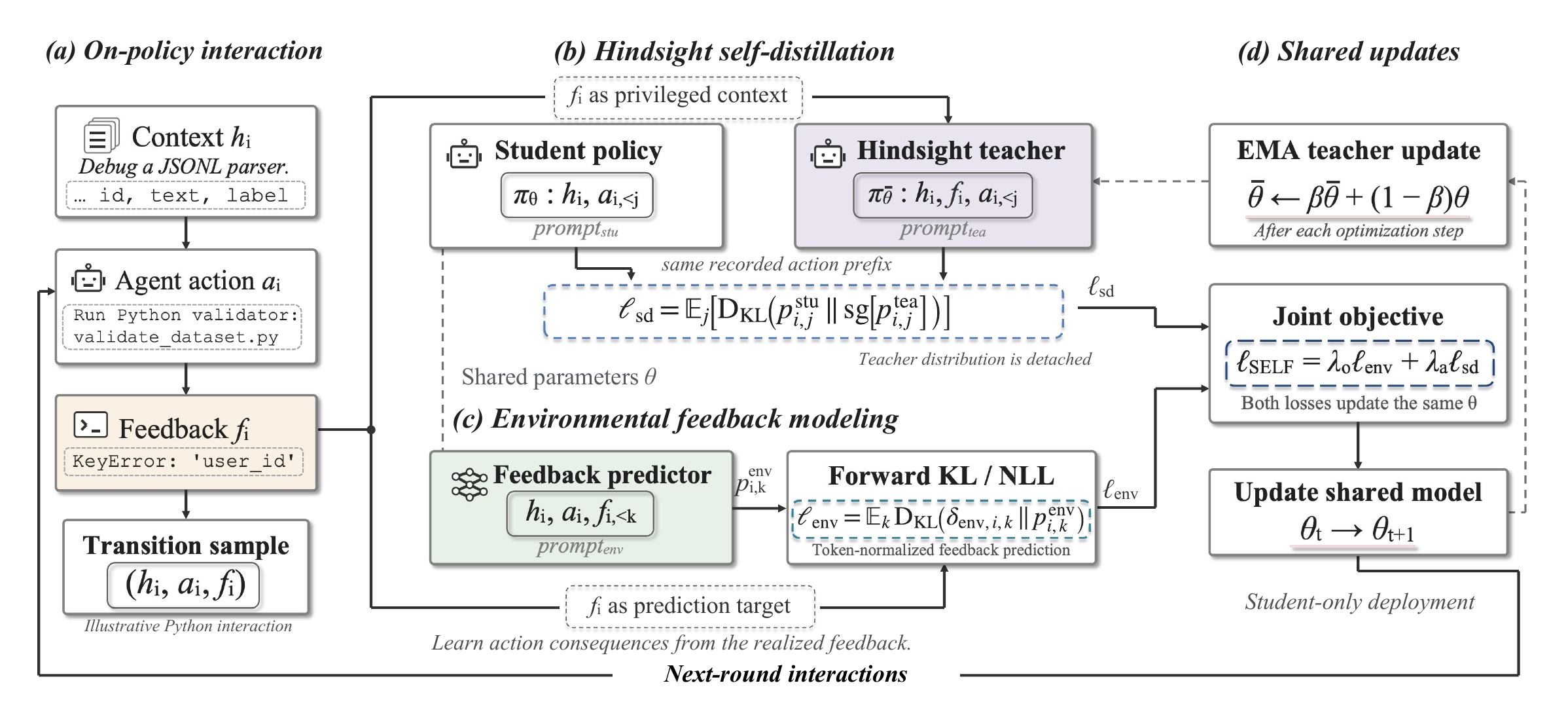}
   \vspace{-15pt}
    \caption{
        Overview of SELF.
        The current policy interacts with the environment to collect
        transitions $(h_i,a_i,f_i)$.
        Each transition is used for two complementary objectives:
        hindsight self-distillation, where a feedback-conditioned
        EMA teacher supervises the student policy, and environmental
        feedback modeling, where the model predicts the realized
        feedback from the interaction history and action.
        The two objectives are jointly optimized to obtain the
        next-round policy. 
    }
    \label{fig:overview}
\end{figure}

\subsection{Problem Formulation}
\label{sec:problem}
\noindent\textbf{Agent--Environment Interaction.}
We consider an agent that interacts with an environment through
actions and observable feedback, including tool outputs, code
execution results, and state changes.
Let $q$ denote a task instruction, $o_0$ the initial observation,
and $f_i$ the feedback returned after action $a_i$.
Before taking $a_i$, the agent has access to the interaction history
\begin{equation}
    h_i=(q,o_0,a_1,f_1,\ldots,a_{i-1},f_{i-1}).
    \label{eq:interaction_history}
\end{equation}
At training round $t$, the current policy and environment generate
\begin{equation}
    a_i \sim \pi_{\theta_t}
    (\cdot\mid h_i,\mathrm{prompt}_{\mathrm{stu}}),
    \;
    f_i \sim \mathcal{P}_{\mathrm{env}}
    (\cdot\mid h_i,a_i),
    \label{eq:agent_interaction}
\end{equation}
where $\pi_{\theta_t}$ is the policy with parameters $\theta_t$,
$\mathrm{prompt}_{\mathrm{stu}}$ specifies the action-generation
interface, and $\mathcal{P}_{\mathrm{env}}$ is the conditional
distribution of environmental feedback.
Each transition $(h_i,a_i,f_i)$  associates an action with
its observed consequence, without requiring access to latent
environment states.

A trajectory is
$\tau=(q,o_0,a_1,f_1,\ldots,a_{n_\tau},f_{n_\tau})$,
where $n_\tau$ is the number of interaction turns, and
$\mathcal{T}_t$ denotes the trajectories collected at round $t$.
Training uses only the resulting actions and environmental feedback,
without task rewards, outcome labels, or external teachers.
For feedback modeling, the observed feedback $f_i$ is represented by
the point-mass target
$\delta_{\mathrm{env},i}(f)=\mathbf{1}\{f=f_i\}$.

\subsection{Hindsight Self-Distillation}
\label{sec:hindsight_self_distillation}

Hindsight self-distillation provides a general approach
for converting environmental feedback into supervision
for policy learning.
We consider its on-policy form, where trajectories are
collected by the current policy and distillation is performed
on the action prefixes encountered along these trajectories~\citep{agarwal2024onpolicy}.
For each transition $(h_i,a_i,f_i)$, the student predicts
the action using only the information available before
the environmental feedback is observed, while a self-teacher
additionally conditions on the realized feedback $f_i$
as privileged hindsight information.
The student and teacher use distinct prompting interfaces,
$\mathrm{prompt}_{\mathrm{stu}}$ and
$\mathrm{prompt}_{\mathrm{tea}}$, respectively.

For an action $a_i$ containing $L_i$ tokens, we define
the student and teacher next-token distributions as
\begin{equation}
\begin{aligned}
    p^{\mathrm{stu}}_{i,j}(\cdot)
    =
    \pi_{\theta}
    \bigl(
        \cdot
        \mid
        h_i,
        a_{i,<j},
        \mathrm{prompt}_{\mathrm{stu}}
    \bigr),
    % \\
    \;
    p^{\mathrm{tea}}_{i,j}(\cdot)
    =
    \pi_{\bar{\theta}}
    \bigl(
        \cdot
        \mid
        h_i,
        f_i,
        a_{i,<j},
        \mathrm{prompt}_{\mathrm{tea}}
    \bigr),
\end{aligned}
\label{eq:hindsight_distribution}
\end{equation}
where both distributions are evaluated on the same
on-policy action prefix $a_{i,<j}$, but only the teacher
has access to the realized environmental feedback.
The teacher parameters $\bar{\theta}$ are maintained
as an exponential moving average (EMA)~\citep{tarvainen2017mean} of the student parameters,
$\bar{\theta}\leftarrow
\beta\bar{\theta}+(1-\beta)\theta$, where $\beta$ is the EMA decay coefficient.

The feedback-conditioned teacher is distilled into
the student through the reverse-KL objective~\citep{gu2024minillm}
\begin{equation}
    \ell_{\mathrm{sd}}
    (h_i,a_i,f_i;\theta,\bar{\theta})
    =
    \mathbb{E}_{j\sim\mathrm{Unif}([L_i])}
    \left[
        D_{\mathrm{KL}}\!\left(
            p^{\mathrm{stu}}_{i,j}
            \,\middle\|\,
            \operatorname{sg}
            [p^{\mathrm{tea}}_{i,j}]
        \right)
    \right],
    \label{eq:action_loss}
\end{equation}
where $[L_i]=\{1,\ldots,L_i\}$ and
$\operatorname{sg}$ stops gradients through the teacher.
The collected action therefore determines the contexts
at which the two distributions are compared, rather than
serving as a ground-truth action target.
This formulation allows information available only after
an action is executed to supervise a policy that must act
without access to that information at inference time~\citep{choudhury2025better}.
In our analysis, however, hindsight self-distillation in
this form yields only modest policy improvements, suggesting
that conditioning the teacher on feedback alone does not
fully exploit the information contained in environmental
feedback.

\begin{accountbox}
\textbf{Why Environmental Feedback Modeling?}
Hindsight self-distillation uses environmental feedback as
privileged context for constructing policy supervision, but
does not explicitly learn the information contained in the
feedback itself.
Our analysis suggests that this limits how effectively the
self-teacher can exploit environmental feedback.

Environmental feedback, however, provides rich supervision
about how the environment responds to an agent's actions.
Each interaction therefore naturally provides a supervised
action--feedback relation,
\[
    (h_i,a_i) \longrightarrow f_i .
\]
This motivates explicitly modeling environmental feedback
alongside hindsight self-distillation, so that the model
learns how the environment responds while using the same
feedback to improve its policy.
\end{accountbox}

\subsection{
Environmental Feedback Modeling}
\label{sec:self_optimization}

Motivated by the observation in the previous section, we treat each transition
$(h_i, a_i, f_i)$ not only as privileged context for hindsight self-distillation,
but also as direct supervision for modeling how the environment responds to an
action. We therefore jointly train environmental feedback modeling and agentic
self-distillation.

Given $(h_i,a_i,f_i)$, we approximate the conditional
feedback distribution
$\mathcal{P}_{\mathrm{env}}(\cdot\mid h_i,a_i)$
with
$p_{\mathrm{env},\theta}
(\cdot\mid h_i,a_i,\mathrm{prompt}_{\mathrm{env}})$,
where $\mathrm{prompt}_{\mathrm{env}}$ is a dedicated
prompt for environmental feedback prediction.
The conditioning context contains the complete interaction
history up to and including $a_i$, while the realized
feedback $f_i$ serves as the prediction target.
This prediction interface shares parameters
$\theta$ with the student policy, rather than introducing
a separate environment model.

For a feedback sequence $f_i$ containing $M_i$ tokens,
let $\delta_{\mathrm{env},i,k}$ denote the point-mass
target on the observed token $f_{i,k}$.
We define the autoregressive next-token distribution as
\begin{equation}
    p^{\mathrm{env}}_{i,k}(\cdot)
    =
    p_{\mathrm{env},\theta}
    \bigl(
        \cdot\mid
        h_i,a_i,f_{i,<k},
        \mathrm{prompt}_{\mathrm{env}}
    \bigr),
    \label{eq:environment_token_distribution}
\end{equation}
where $f_{i,<k}$ denotes the observed feedback prefix.
The environmental feedback modeling objective is
\begin{equation}
    \ell_{\mathrm{env}}(h_i,a_i,f_i;\theta)
    =
    \mathbb{E}_{k\sim\mathrm{Unif}([M_i])}
    \left[
        D_{\mathrm{KL}}\!\left(
            \delta_{\mathrm{env},i,k}
            \,\middle\|\,
            p^{\mathrm{env}}_{i,k}
        \right)
    \right].
    \label{eq:observation_loss}
\end{equation}
Because each target is a point mass, this forward-KL
objective is equivalent to the token-normalized
negative log-likelihood of the realized feedback.
It therefore trains the shared model to capture observable
action--feedback relationships directly from interaction,
without requiring latent environment states or treating
the sampled action as a correct action target.

\begin{table}[t]
\caption{
Comparison of learning paradigms under the configurations
considered in this work.
Supervision describes the granularity of the learning signal.
Teacher denotes the source of teacher supervision.
Env.\ Modeling indicates whether an explicit objective predicts
environmental responses from interaction histories and actions.
Dashes indicate inapplicable attributes.
}
\label{tab:learning-paradigms}
\begin{center}
\small
\setlength{\tabcolsep}{5pt}
\renewcommand{\arraystretch}{1.15}
\begin{tabular}{@{}llllll@{}}
\toprule
\textbf{Method}
& \textbf{Sampling}
& \textbf{Supervision}
& \shortstack[l]{\textbf{Feedback}\\\textbf{Source}}
& \textbf{Teacher}
& \shortstack[l]{\textbf{Env.}\\\textbf{Modeling}} \\
\midrule
SFT  & Off-policy & Dense  & Teacher     & External & No \\
GRPO & On-policy  & Sparse & Verifier    & None     & No \\
Mem0 & --         & --     & Environment & None     & No \\
SDPO & On-policy  & Dense  & Environment & Self     & No \\
\rowcolor{selfbg}
\textbf{SELF (Ours)}
     & On-policy  & Dense  & Environment & Self     & Yes \\
\bottomrule
\end{tabular}
\end{center}
\end{table}

\subsection{Overall SELF Training Objective}
The proposed \textbf{SELF} framework jointly
optimizes environmental feedback modeling and hindsight
self-distillation within a shared model.
The same realized feedback $f_i$ serves two complementary roles:
it provides a prediction target for learning how the environment
responds to an action, while also acting as privileged hindsight
information for constructing policy supervision.
By optimizing both objectives through the same underlying model,
SELF jointly improves environmental feedback modeling and
agent action generation.
For each transition $(h_i,a_i,f_i)$, the SELF objective is
\begin{equation}
    \ell_{\mathrm{SELF}}
    =
    \lambda_o \ell_{\mathrm{env}}
    +
    \lambda_a \ell_{\mathrm{sd}},
    \label{eq:self_transition_loss}
\end{equation}
where $\lambda_o$ and $\lambda_a$ control the contributions
of environmental feedback modeling and hindsight
self-distillation, respectively.

At training round $t$, the current policy collects a set of
trajectories $\mathcal{T}_t$. Averaging the token-normalized
objectives over trajectories and interaction turns yields
\begin{equation}
\begin{aligned}
\mathcal{L}_{\mathrm{SELF},t}(\theta)
=
\mathbb{E}_{\tau\sim\mathrm{Unif}(\mathcal{T}_t)}
\mathbb{E}_{i\sim\mathrm{Unif}([n_\tau])}
\Bigg[
&
\lambda_o
\mathbb{E}_{k\sim\mathrm{Unif}([M_i])}
D_{\mathrm{KL}}\!\left(
    \delta_{\mathrm{env},i,k}
    \,\middle\|\,
    p^{\mathrm{env}}_{i,k}
\right)
\\
&+
\lambda_a
\mathbb{E}_{j\sim\mathrm{Unif}([L_i])}
D_{\mathrm{KL}}\!\left(
    p^{\mathrm{stu}}_{i,j}
    \,\middle\|\,
    \operatorname{sg}
    [p^{\mathrm{tea}}_{i,j}]
\right)
\Bigg],
\end{aligned}
\label{eq:self_objective}
\end{equation}
where $n_\tau$ denotes the number of interaction turns
in trajectory $\tau$.
% The outer expectations balance trajectories and interaction
% turns, while the inner expectations normalize the two
% learning objectives over their respective token sequences.
Optimizing $\mathcal{L}_{\mathrm{SELF},t}$ updates the shared
parameters from $\theta_t$ to $\theta_{t+1}$.
Environmental feedback modeling therefore updates the same
model from which subsequent hindsight supervision is derived,
while self-distillation directly improves the deployable
policy and simultaneously updates the representation used
for feedback prediction.
The updated policy then collects new interactions for the
next training round.
At deployment, only the student policy
$\pi_{\theta}(\cdot\mid h_i,\mathrm{prompt}_{\mathrm{stu}})$
is required.
The complete training procedure is summarized in
Algorithm~\ref{alg:self} in the appendix, and illustrative examples of
$\mathrm{prompt}_{\mathrm{stu}}$,
$\mathrm{prompt}_{\mathrm{tea}}$, and
$\mathrm{prompt}_{\mathrm{env}}$ are provided in the appendix.

\section{Experiments}
\label{sec:evaluation}

In this section, we evaluate SELF on $\tau$-bench Retail and AppWorld
and address the following three research questions (RQs). \noindent\textbf{RQ1:} Can SELF improve interactive agent performance using
environmental feedback without verifiable rewards?
\noindent\textbf{RQ2:} Do environmental feedback modeling and hindsight
self-distillation mutually improve each other?
\noindent\textbf{RQ3:} Does SELF retain general language-model capabilities
after agent training?

\begin{table}[t]
    \caption{
        Performance on $\tau$-bench and AppWorld.
        SR denotes the $\mathrm{pass}@4$ success rate on $\tau$-bench,
        and Turns denotes the average number of interaction turns.
        TGC and SGC denote task goal completion and scenario goal
        completion on AppWorld, respectively.
        Observation and Full denote baseline feedback settings,
        while Reward denotes reward-based training.
        SELF learns from agent-generated actions and environmental
        observations, and its rows are shaded light blue.
        Higher SR, TGC, and SGC and lower Turns are better.
        Within each backbone, excluding external model baselines,
        the best means are \textbf{bold} with pale gold shading,
        and the second-best means are \underline{underlined}
        with pale pink shading.
        Missing results and inapplicable settings are denoted by --.
    }
    \label{tab:main_results}

    \begin{center}
        \small
        \setlength{\tabcolsep}{5pt}
        \renewcommand{\arraystretch}{1.12}

        \begin{tabular}{@{}llcccc@{}}
            \toprule
            \multirow{2}{*}{\textbf{Method}}
            & \multirow{2}{*}{\textbf{Setting}}
            & \multicolumn{2}{c}{\textbf{$\tau$-bench}}
            & \multicolumn{2}{c}{\textbf{AppWorld}} \\
            \cmidrule(lr){3-4}
            \cmidrule(lr){5-6}
            & &
            SR $\uparrow$
            & Turns $\downarrow$
            & TGC $\uparrow$
            & SGC $\uparrow$ \\
            \midrule

            \multicolumn{6}{c}{
                \textbf{\textit{External Model Baselines}}
            } \\
            \midrule

            Qwen3-30B-A3B-Instruct
            & -- & 33.6 & 11.9 & 31.55 & 12.50 \\

            \midrule
            \multicolumn{6}{c}{
                \textbf{\textit{Qwen3-4B-Instruct}}
            } \\
            \midrule

            Base
            & --
            & 27.2 $\pm$ 1.4
            & \cellcolor{secondbg}\underline{10.5} $\pm$ 0.3
            & 16.67 $\pm$ 1.19
            & 5.36 $\pm$ 0.62 \\

            OEL
            & Observation
            & 30.6 $\pm$ 1.5
            & 10.9 $\pm$ 0.4
            & 19.05 $\pm$ 1.43
            & 6.25 $\pm$ 0.74 \\

            SDPO
            & Observation
            & 30.4 $\pm$ 1.7
            & 11.5 $\pm$ 0.5
            & 20.24 $\pm$ 1.56
            & 7.14 $\pm$ 0.83 \\

            Mem0
            & Observation
            & 19.2 $\pm$ 1.8
            & 12.6 $\pm$ 0.6
            & 11.90 $\pm$ 1.31
            & 3.57 $\pm$ 0.58 \\

            SDPO
            & Full
            & 22.5 $\pm$ 1.6
            & 12.4 $\pm$ 0.5
            & 14.29 $\pm$ 1.47
            & 4.46 $\pm$ 0.69 \\

            Mem0
            & Full
            & 27.4 $\pm$ 1.5
            & 11.7 $\pm$ 0.4
            & 17.86 $\pm$ 1.38
            & 5.95 $\pm$ 0.77 \\

            GRPO
            & Reward
            & \cellcolor{secondbg}\underline{33.3} $\pm$ 1.9
            & 13.4 $\pm$ 0.7
            & \cellcolor{secondbg}\underline{28.57} $\pm$ 1.72
            & \cellcolor{secondbg}\underline{10.71} $\pm$ 1.04 \\

            % \rowcolor{carow}
            \rowcolor{selfbg}
            % \textbf{SELF}
            \textcolor{selfaccent}{\textbf{SELF (Ours)}}
            & Observation
            & \cellcolor{bestbg}\textbf{38.7} $\pm$ 1.5
            & \cellcolor{bestbg}\textbf{10.4} $\pm$ 0.4
            & \cellcolor{bestbg}\textbf{32.14} $\pm$ 1.61
            & \cellcolor{bestbg}\textbf{12.50} $\pm$ 0.96 \\

            \midrule
            \multicolumn{6}{c}{
                \textbf{\textit{Qwen3-8B (Non-Thinking)}}
            } \\
            \midrule

            Base
            & --
            & 35.7 $\pm$ 1.3
            & \cellcolor{bestbg}\textbf{10.0} $\pm$ 0.3
            & 22.62 $\pm$ 1.28
            & 7.14 $\pm$ 0.71 \\

            OEL
            & Observation
            & 37.9 $\pm$ 1.4
            & \cellcolor{secondbg}\underline{10.4} $\pm$ 0.4
            & 26.19 $\pm$ 1.46
            & 8.04 $\pm$ 0.82 \\

            SDPO
            & Observation
            & 36.8 $\pm$ 1.6
            & 11.0 $\pm$ 0.4
            & 25.00 $\pm$ 1.53
            & 8.93 $\pm$ 0.89 \\

            Mem0
            & Observation
            & 29.5 $\pm$ 1.7
            & 12.0 $\pm$ 0.6
            & 16.67 $\pm$ 1.35
            & 4.36 $\pm$ 0.63 \\

            SDPO
            & Full
            & 30.6 $\pm$ 1.5
            & 11.8 $\pm$ 0.5
            & 19.05 $\pm$ 1.42
            & 6.25 $\pm$ 0.76 \\

            Mem0
            & Full
            & 35.8 $\pm$ 1.4
            & 11.1 $\pm$ 0.4
            & 22.62 $\pm$ 1.39
            & 7.14 $\pm$ 0.84 \\

            GRPO
            & Reward
            & \cellcolor{secondbg}\underline{39.1} $\pm$ 1.8
            & 12.7 $\pm$ 0.6
            & \cellcolor{secondbg}\underline{32.14} $\pm$ 1.68
            & \cellcolor{secondbg}\underline{11.61} $\pm$ 1.02 \\

            % \rowcolor{carow}
            \rowcolor{selfbg}
            % \textbf{SELF}
            \textcolor{selfaccent}{\textbf{SELF (Ours)}}
            & Observation
            & \cellcolor{bestbg}\textbf{43.2} $\pm$ 1.4
            & 10.8 $\pm$ 0.4
            & \cellcolor{bestbg}\textbf{35.71} $\pm$ 1.57
            & \cellcolor{bestbg}\textbf{13.39} $\pm$ 0.93 \\

            \bottomrule
        \end{tabular}
    \end{center}
\end{table}

\subsection{Experimental Setup}
\label{sec:experimental_setup}

\textbf{Benchmark Datasets.}
We evaluate SELF on $\tau$-bench Retail~\citep{yao2025taubench} and AppWorld~\citep{trivedi2024appworld},
covering multi-turn customer-service interactions with
domain-specific tools and code-based task completion
across applications, respectively.
We use the official training splits for learning and
held-out test splits for evaluation.
On $\tau$-bench Retail, we report success rate
(SR; $\mathrm{pass}@4$) and the average number of
interaction turns (Turns) to assess task performance
and interaction efficiency.
On AppWorld, we report Task Goal Completion (TGC)
and Scenario Goal Completion (SGC), measuring task-level
success and consistency across task variants, respectively.
We additionally evaluate MMLU~\citep{hendrycks2021mmlu}, MMLU-Pro~\citep{wang2024mmlupro}, and IFEval~\citep{zhou2023ifeval}
to assess the retention of general capabilities
after agent training.

\textbf{Comparing Baselines.}
We compare SELF with hindsight-distillation methods
OEL~\citep{ye2026oel} and SDPO~\citep{hubotter2026sdpo},
the memory-based method Mem0~\citep{chhikara2025mem0},
and the reinforcement learning method
GRPO~\citep{shao2024deepseekmath}.
We distinguish three supervision settings:
\textit{Observation} supplies only environmental observations,
without ground-truth task outcomes or success/failure labels;
\textit{Full} additionally supplies ground-truth information
and explicit success/failure feedback; and \textit{Reward}
provides verifiable task rewards for reinforcement learning.
OEL and SELF use Observation, SDPO and Mem0 are evaluated
under both Observation and Full, and GRPO uses Reward.
Table~\ref{tab:learning-paradigms} summarizes the methods'
sampling strategies, supervision sources, and use of
environment modeling. We also report unadapted base models and a strong
external model to contextualize trained-agent performance:
Qwen3-30B-A3B-Instruct~\citep{yang2025qwen3}.

\textbf{Implementation Details.}
Our main experiments use Qwen3-4B-Instruct and
Qwen3-8B (non-thinking) as backbone models~\citep{yang2025qwen3}.
All ablation studies, mechanism analyses, and
general-capability evaluations use Qwen3-4B-Instruct.
Models evaluated in our framework operate under
the same ReAct-style tool-use interface~\citep{yao2023react}.
For SELF, the self-distillation teacher is maintained
as an exponential moving average (EMA)~\citep{tarvainen2017mean} of the student,
with decay coefficient $\beta=0.99$.
Unless otherwise specified, we use $\lambda_o=0.2$
and $\lambda_a=1.0$ as the default loss weights.
We repeat each experiment over three independent runs
and report the mean together with the standard deviation.
Complete training hyperparameters and implementation
details are provided in the appendix.

\begin{figure}[t]
    \centering
    \includegraphics[width=\linewidth]{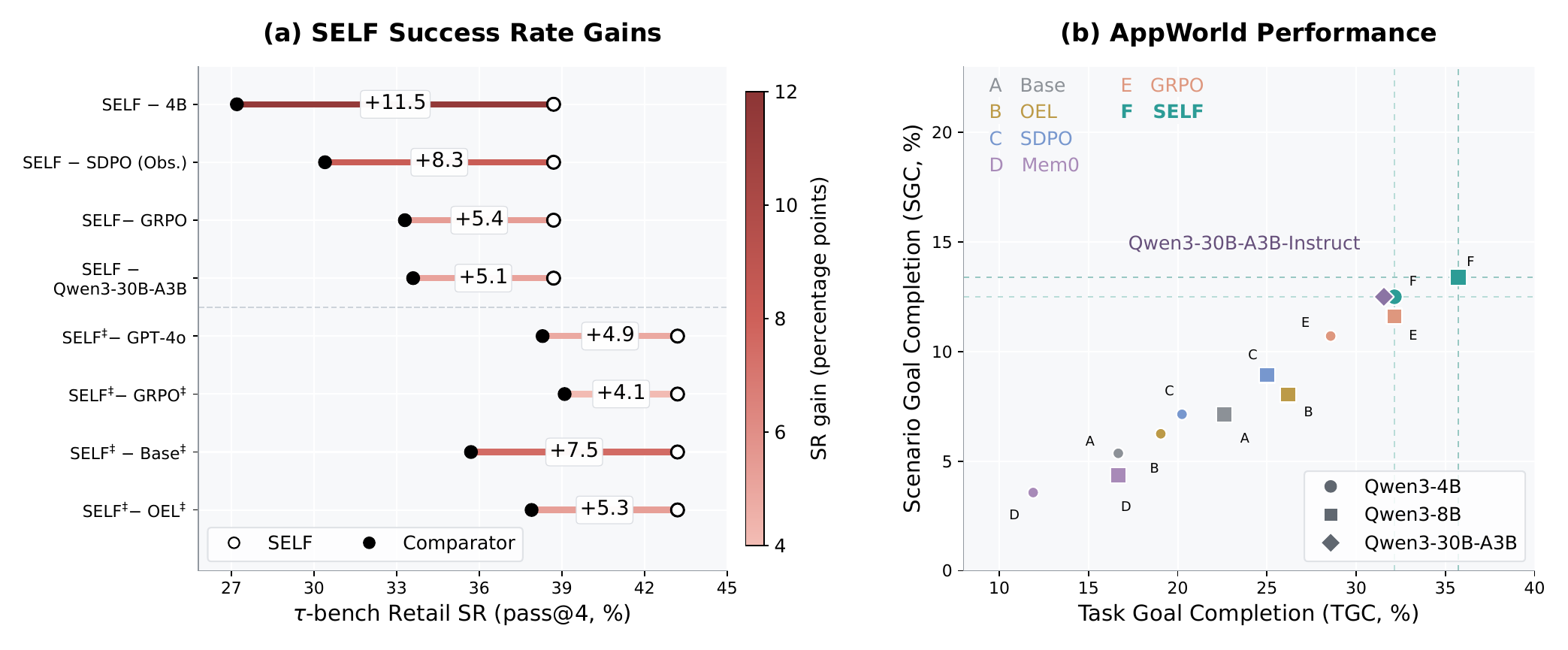}
    \vspace{-15pt}
    \caption{
        Performance comparison on $\tau$-bench Retail and AppWorld.
        (a) Success rate (SR, pass@4) gains of SELF over selected
        comparators on $\tau$-bench Retail.
        Open and filled circles denote SELF and comparator scores,
        respectively; annotations and line colors indicate gains
        in percentage points.
        The superscript $\ddagger$ denotes Qwen3-8B;
        unmarked methods use Qwen3-4B-Instruct unless otherwise specified.
        (b) Task goal completion (TGC) and scenario goal completion
        (SGC) on AppWorld. Colors distinguish methods, and marker
        shapes distinguish model backbones.
    }
    \label{fig:main_comparison}
\end{figure}

\subsection{Main Results (RQ1)}

We first evaluate whether SELF improves interactive agent performance
using environmental feedback alone, comparing it with self-distillation,
memory-based, and reward-trained baselines under different supervision
settings. Table~\ref{tab:main_results} and Figure~\ref{fig:main_comparison}
report task performance and interaction efficiency on $\tau$-bench Retail
and AppWorld using Qwen3-4B-Instruct and Qwen3-8B (non-thinking).
Under Observation, SELF consistently outperforms SDPO and OEL
on SR, TGC, and SGC. With Qwen3-4B-Instruct, SELF achieves
38.7\% SR, exceeding SDPO and OEL by 8.3 and 8.1 percentage
points; the corresponding gains with Qwen3-8B are 6.4 and
5.3 points. These results indicate more effective use of
environmental feedback than existing self-distillation methods.
SELF also surpasses Mem0 under both Observation and Full:
with Qwen3-4B-Instruct, Mem0 achieves 19.2\% and 27.4\% SR,
respectively, versus SELF's 38.7\%.
SELF further outperforms SDPO under Full feedback.
Indeed, Full feedback reduces SDPO's SR from 30.4\% to 22.5\%
with Qwen3-4B-Instruct and from 36.8\% to 30.6\% with Qwen3-8B,
alongside lower AppWorld TGC and SGC.
Additional ground-truth information and success/failure labels
therefore do not necessarily improve feedback-based learning.
Finally, SELF exceeds reward-trained GRPO by 5.4 and 4.1 SR
points on the two backbones, respectively, while also achieving
higher TGC and SGC. Together, these comparisons demonstrate
effective policy learning from environmental feedback without verifiable rewards.

\subsection{Mutual Benefits of Feedback Modeling and Self-Distillation (RQ2)}
We examine the interaction between the two objectives in SELF.
Specifically, we test whether environmental feedback modeling improves
feedback-derived hindsight supervision, and
whether hindsight self-distillation improves the shared model's ability
to predict environmental feedback.

\begin{figure}[t]
    \centering
    \includegraphics[width=\linewidth]{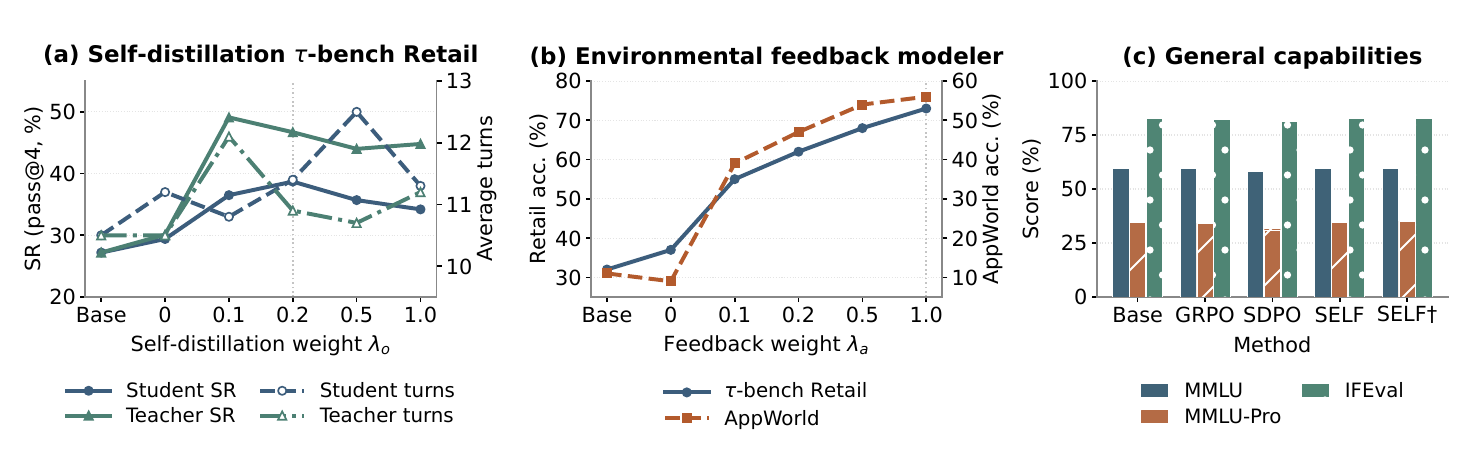}
    \vspace{-15pt}
    \caption{
    Ablation studies and general capability evaluation
    with Qwen3-4B-Instruct.
    (a) Student and privileged teacher success rates
    (SR, pass@4) and average interaction turns on
    $\tau$-bench Retail across different values of $\lambda_o$.
    (b) Environmental feedback prediction accuracy on
    $\tau$-bench Retail and AppWorld across different
    values of $\lambda_a$.
    (c) Performance on MMLU, MMLU-Pro, and IFEval;
    SELF${\dagger}$ denotes the privileged teacher.
    Vertical dashed lines in (a) and (b) indicate default
    settings. }
    \label{fig:ablation-lambda}
\end{figure}

\begin{table}[h]
    \caption{
        Effect of the environmental feedback modeling weight $\lambda_o$
        on $\tau$-bench Retail with Qwen3-4B-Instruct.
        SR is reported as pass@4 (\%), and Turns denotes
        the average number of interaction turns.
        \textbf{Bold} indicates the best result in each column.
    }
    \label{tab:lambda_o_retail}

    \begin{center}
        \small
        \setlength{\tabcolsep}{6pt}
        \renewcommand{\arraystretch}{1.1}

        \begin{tabular}{lcccc}
            \toprule
            \multirow{2}{*}{Setting}
            & \multicolumn{2}{c}{Student}
            & \multicolumn{2}{c}{Privileged teacher} \\
            \cmidrule(lr){2-3}
            \cmidrule(lr){4-5}
            & SR $\uparrow$
            & Turns $\downarrow$
            & SR $\uparrow$
            & Turns $\downarrow$ \\
            \midrule

            Base
            & 27.2 $\pm$ 1.4
            & 10.5 $\pm$ 0.3
            & -- & -- \\
            \midrule

            $\lambda_o = 0$
            & 29.4 $\pm$ 2.3
            & 11.2 $\pm$ 0.5
            & 30.1 $\pm$ 1.1
            & \textbf{10.5} $\pm$ 0.2 \\

            $\lambda_o = 0.1$
            & 36.5 $\pm$ 0.9
            & 10.8 $\pm$ 0.4
            & \textbf{49.1} $\pm$ 1.8
            & 12.1 $\pm$ 0.6 \\

            \rowcolor{selfbg}
            $\lambda_o = 0.2$
            & \textbf{38.7} $\pm$ 1.5
            & \textbf{10.4} $\pm$ 0.4
            & 46.7 $\pm$ 2.4
            & 10.9 $\pm$ 0.4 \\

            $\lambda_o = 0.5$
            & 35.7 $\pm$ 2.1
            & 12.5 $\pm$ 0.6
            & 44.0 $\pm$ 1.3
            & 10.7 $\pm$ 0.3 \\

            $\lambda_o = 1.0$
            & 34.2 $\pm$ 1.2
            & 11.3 $\pm$ 0.3
            & 44.8 $\pm$ 2.0
            & 11.2 $\pm$ 0.5 \\

            \bottomrule
        \end{tabular}
    \end{center}
\end{table}

\textbf{Environmental feedback modeling strengthens the hindsight
teacher and improves feedback-derived supervision.}
We vary the observation-modeling weight
$\lambda_o \in \{0, 0.1, 0.2, 0.5, 1.0\}$
while fixing the self-distillation weight, using
Qwen3-4B-Instruct on $\tau$-bench Retail.
Table~\ref{tab:lambda_o_retail} and
Figure~\ref{fig:ablation-lambda}(a) report student and
privileged-teacher performance; the teacher evaluation
protocol is detailed in the appendix.
At the default $\lambda_o=0.2$, the student achieves
its highest SR of 38.7\%, exceeding Base by 11.5 percentage
points, while the teacher reaches 46.7\%.
The teacher peaks at 49.1\% with $\lambda_o=0.1$,
but the corresponding student achieves only 36.5\%,
showing that higher privileged-teacher performance
does not necessarily produce the best distilled student.
Compared with no observation modeling ($\lambda_o=0$),
the default improves teacher SR from 30.1\% to 46.7\%
and student SR from 29.4\% to 38.7\%, gains of 16.6
and 9.3 points, respectively.
These results support the interpretation that feedback modeling strengthens the feedback-conditioned teacher and its ability to provide effective policy supervision, with the student gains demonstrating benefits for hindsight self-distillation, even when learning relies on observed environmental responses.

\begin{table}[t]
    \caption{
        Effect of the self-distillation weight $\lambda_a$
        on $\tau$-bench Retail and AppWorld with Qwen3-4B-Instruct.
        SR is reported as pass@4, and Turns denotes the average
        number of interaction turns.
        All metrics except Turns are reported as percentages.
    }
    \label{tab:lambda_a_ablation}

    \begin{center}
        \small
        \setlength{\tabcolsep}{6pt}
        \renewcommand{\arraystretch}{1.1}

        \begin{tabular}{lcccccc}
            \toprule
            \multirow{2}{*}{Setting}
            & \multicolumn{3}{c}{$\tau$-bench Retail}
            & \multicolumn{3}{c}{AppWorld} \\
            \cmidrule(lr){2-4}
            \cmidrule(lr){5-7}
            & SR $\uparrow$
            & Turns $\downarrow$
            & Acc. $\uparrow$
            & TGC $\uparrow$
            & SGC $\uparrow$
            & Acc. $\uparrow$ \\
            \midrule

            Base
            & 27.2 $\pm$ 1.4
            & 10.5 $\pm$ 0.3
            & 32 $\pm$ 1.8
            & 16.67 $\pm$ 1.19
            & 5.36 $\pm$ 0.62
            & 11 $\pm$ 1.3 \\
            \midrule

            $\lambda_a = 0$
            & 20.1 $\pm$ 2.1
            & 11.2 $\pm$ 0.5
            & 37 $\pm$ 2.6
            & 5.52 $\pm$ 0.83
            & 1.21 $\pm$ 0.34
            & 9 $\pm$ 1.7 \\

            $\lambda_a = 0.1$
            & 31.5 $\pm$ 1.3
            & 12.0 $\pm$ 0.6
            & 55 $\pm$ 1.5
            & 26.67 $\pm$ 2.07
            & 8.62 $\pm$ 0.91
            & 39 $\pm$ 2.4 \\

            $\lambda_a = 0.2$
            & 33.2 $\pm$ 1.8
            & 10.7 $\pm$ 0.3
            & 62 $\pm$ 2.2
            & 29.91 $\pm$ 1.43
            & 8.45 $\pm$ 1.16
            & 47 $\pm$ 1.6 \\

            $\lambda_a = 0.5$
            & 36.5 $\pm$ 0.9
            & \textbf{10.2} $\pm$ 0.2
            & 68 $\pm$ 1.3
            & 31.10 $\pm$ 1.86
            & 9.73 $\pm$ 0.78
            & 54 $\pm$ 2.1 \\

            \rowcolor{selfbg}
            $\lambda_a = 1.0$
            & \textbf{38.7} $\pm$ 1.5
            & 10.4 $\pm$ 0.4
            & \textbf{73} $\pm$ 1.7
            & \textbf{32.14} $\pm$ 1.61
            & \textbf{12.50} $\pm$ 0.96
            & \textbf{56} $\pm$ 1.2 \\

            \bottomrule
        \end{tabular}
    \end{center}
\end{table}

\textbf{Self-distillation strengthens environmental feedback modeling,
supporting the reverse direction of their mutually reinforcing relationship.}
We vary the self-distillation weight
$\lambda_a \in \{0, 0.1, 0.2, 0.5, 1.0\}$
while fixing $\lambda_o$, using Qwen3-4B-Instruct on
$\tau$-bench Retail and AppWorld.
We measure feedback prediction accuracy (Acc.) on a shared,
fixed set of held-out transitions collected by a reference policy
in each environment.
All variants receive identical histories and actions and are
evaluated against the same observed feedback under identical
evaluation settings
(see appendix for details).
As shown in Table~\ref{tab:lambda_a_ablation} and
Figure~\ref{fig:ablation-lambda}(b), accuracy increases
monotonically across the tested $\lambda_a$ values,
peaking at $\lambda_a=1.0$: 73\% on Retail and 56\% on
AppWorld, compared with 37\% and 9\% without self-distillation.
Policy performance improves alongside prediction accuracy:
increasing $\lambda_a$ from 0 to 1.0 raises Retail SR
from 20.1\% to 38.7\% and AppWorld TGC from 5.52\% to 32.14\%.
Without self-distillation, task performance falls below Base
on both benchmarks, highlighting the importance of the
policy-learning objective.
Together with the $\lambda_o$ ablation, results support
a mutually reinforcing relationship: feedback modeling
improves feedback-derived policy supervision,
while hindsight self-distillation enhances the shared model's
ability to predict action consequences.
Feedback modeling benefits from policy learning while supporting it.

\begin{table}[t]
    \caption{
        General capability evaluation on MMLU, MMLU-Pro,
        and IFEval using Qwen3-4B-Instruct.
        Results are mean $\pm$ standard deviation.
        Light and darker blue shading identify the SELF student
        and teacher, respectively.
        Higher is better; \textbf{bold} entries indicate
        the highest mean per column.
    }
    \label{tab:general_capabilities}

    \begin{center}
        \small
        \setlength{\tabcolsep}{6pt}
        \renewcommand{\arraystretch}{1.12}

        \begin{tabular}{@{}llccc@{}}
            \toprule
            \textbf{Method}
            & \textbf{Setting}
            & \textbf{MMLU} $\uparrow$
            & \textbf{MMLU-Pro} $\uparrow$
            & \textbf{IFEval} $\uparrow$ \\
            \midrule

            Base
            & --
            & 59.60 $\pm$ 0.18
            & 34.38 $\pm$ 0.27
            & \textbf{82.62} $\pm$ 0.35 \\

            GRPO
            & Reward
            & 59.61 $\pm$ 0.24
            & 34.25 $\pm$ 0.36
            & 82.19 $\pm$ 0.48 \\

            SDPO
            & Observation
            & 57.96 $\pm$ 0.31
            & 31.02 $\pm$ 0.42
            & 81.17 $\pm$ 0.53 \\

            \rowcolor{studentbg}
            SELF
            & Observation
            & 59.57 $\pm$ 0.22
            & 34.49 $\pm$ 0.33
            & 82.61 $\pm$ 0.41 \\

            \rowcolor{teacherbg}
            SELF (teacher)
            & Observation
            & \textbf{59.71} $\pm$ 0.20
            & \textbf{34.92} $\pm$ 0.29
            & 82.58 $\pm$ 0.38 \\

            \bottomrule
        \end{tabular}
    \end{center}
\end{table}

\subsection{General Capability Retention (RQ3)}

Finally, we examine whether SELF's gains in interactive environments
come at the expense of broader capabilities by evaluating
the Qwen3-4B-Instruct student and EMA teacher on MMLU,
MMLU-Pro, and IFEval, covering general knowledge, reasoning,
and instruction following
(Table~\ref{tab:general_capabilities} and
Figure~\ref{fig:ablation-lambda}).
The SELF student closely matches the base model:
MMLU changes from 59.60 to 59.57, MMLU-Pro from 34.38
to 34.49, and IFEval from 82.62 to 82.61,
with absolute differences of at most 0.11 points.
In contrast, observation-based SDPO reduces these scores
to 57.96, 31.02, and 81.17, respectively.
The EMA teacher similarly retains general performance.
These results support capability retention rather than
general capability improvements, showing that SELF's
interactive-task gains accompany minimal changes on
the evaluated general benchmarks.

\section{Conclusion}
\label{sec:conclusion}

We introduced SELF, a framework that jointly optimizes environmental
feedback modeling and hindsight self-distillation, using realized
action consequences both as prediction targets for learning
action--feedback relationships and as privileged context for
self-teacher supervision.
Experiments on $\tau$-bench Retail and AppWorld show that SELF
consistently outperforms existing feedback-learning methods and
reward-trained GRPO while largely preserving the general capabilities
of the base model. Our analysis further reveals a mutually reinforcing mechanism:
environmental feedback modeling strengthens hindsight self-distillation,
while self-distillation improves feedback prediction.
These findings suggest that effective learning from environmental
feedback requires not only conditioning on the observed feedback,
but also explicitly learning the action--feedback regularities encoded
in interaction. SELF therefore provides a simple framework for jointly learning how
the environment responds and how the agent should act.

% \newpage

\bibliography{references}

@inproceedings{yao2023react,
  title = {{ReAct}: Synergizing Reasoning and Acting in Language Models},
  author = {Shunyu Yao and Jeffrey Zhao and Dian Yu and Nan Du and Izhak Shafran and Karthik Narasimhan and Yuan Cao},
  year = {2023},
  booktitle = {The Eleventh International Conference on Learning Representations},
  url = {https://openreview.net/forum?id=WE_vluYUL-X}
}

@article{song2026rltf,
  title     = {Expanding the Capabilities of Reinforcement Learning via Text Feedback},
  author    = {Song, Yuda and Chen, Lili and Tajwar, Fahim and Munos, Remi
               and Pathak, Deepak and Bagnell, J. Andrew and Singh, Aarti
               and Zanette, Andrea},
  journal   = {arXiv preprint arXiv:2602.02482},
  year      = {2026},
  doi       = {10.48550/arXiv.2602.02482}
}

@inproceedings{zhang2026earlyexperience,
  title     = {Agent Learning via Early Experience},
  author    = {Zhang, Kai and Chen, Xiangchao and Liu, Bo and Xue, Tianci
               and Liao, Zeyi and Liu, Zhihan and Wang, Xiyao and Ning, Yuting
               and Chen, Zhaorun and Fu, Xiaohan and Xie, Jian and Sun, Yuxuan
               and Gou, Boyu and Qi, Qi and Meng, Zihang and Yang, Jianwei
               and Zhang, Ning and Li, Xian and Shah, Ashish and Huynh, Dat
               and Li, Hengduo and Yang, Zi and Cao, Xuefei
               and Jang, Lawrence Keunho and Zhou, Shuyan and Zhu, Jiacheng
               and Sun, Huan and Weston, Jason E. and Su, Yu and Wu, Yifan},
  booktitle = {Forty-third International Conference on Machine Learning},
  year      = {2026}
}

@article{jin2026ahead,
  title     = {AHEAD: Adaptive Hindsight with Environment-Augmented Distillation for Agentic RL},
  author    = {Jin, Xiaolong and Wang, Dingmin and Lingam, Vijay and Kumar, Varun},
  journal   = {arXiv preprint arXiv:2608.24114},
  year      = {2026},
  doi       = {10.48550/arXiv.2608.24114}
}

@article{wang2026openclawrl,
  title     = {OpenClaw-RL: Train Any Agent Simply by Talking},
  author    = {Wang, Yinjie and Chen, Xuyang and Jin, Xiaolong
               and Wang, Mengdi and Yang, Ling},
  journal   = {arXiv preprint arXiv:2603.10165},
  year      = {2026},
  doi       = {10.48550/arXiv.2603.10165}
}

@inproceedings{loshchilov2019decoupled,
  title     = {Decoupled Weight Decay Regularization},
  author    = {Loshchilov, Ilya and Hutter, Frank},
  booktitle = {International Conference on Learning Representations},
  year      = {2019},
  url       = {https://openreview.net/forum?id=Bkg6RiCqY7}
}

@article{yang2026rlsd,
  title     = {Self-Distilled RLVR},
  author    = {Yang, Chenxu and Qin, Chuanyu and Si, Qingyi and Chen, Minghui
               and Gu, Naibin and Yao, Dingyu and Lin, Zheng and Wang, Weiping
               and Wang, Jiaqi and Duan, Nan},
  journal   = {arXiv preprint arXiv:2604.03128},
  year      = {2026},
  doi       = {10.48550/arXiv.2604.03128}
}

@article{buening2026aligning,
  title     = {Aligning Language Models from User Interactions},
  author    = {Kleine Buening, Thomas and H{\"u}botter, Jonas
               and P{\'a}sztor, Barna and Shenfeld, Idan
               and Ramponi, Giorgia and Krause, Andreas},
  journal   = {arXiv preprint arXiv:2603.12273},
  year      = {2026},
  doi       = {10.48550/arXiv.2603.12273}
}

@inproceedings{trivedi2024appworld,
  title = {{AppWorld}: A Controllable World of Apps and People for Benchmarking Interactive Coding Agents},
  author = {Harsh Trivedi and Tushar Khot and Mareike Hartmann and Ruskin Manku and Vinty Dong and Edward Li and Shashank Gupta and Ashish Sabharwal and Niranjan Balasubramanian},
  year = {2024},
  booktitle = {Proceedings of the 62nd Annual Meeting of the Association for Computational Linguistics (Volume 1: Long Papers)},
  pages = {16022--16076},
  publisher = {Association for Computational Linguistics},
  doi = {10.18653/v1/2024.acl-long.850},
  url = {https://aclanthology.org/2024.acl-long.850/}
}

@inproceedings{choudhury2025better,
  title = {Better than Your Teacher: {LLM} Agents that Learn from Privileged {AI} Feedback},
  author = {Sanjiban Choudhury and Paloma Sodhi},
  booktitle = {The Thirteenth International Conference on Learning Representations},
  year = {2025},
  url = {https://proceedings.iclr.cc/paper_files/paper/2025/hash/1c60ed2b01120d383eebf12dc7a0e138-Abstract-Conference.html}
}

@inproceedings{lambert2025tulu3,
  title = {{T{\"u}lu 3}: Pushing Frontiers in Open Language Model Post-Training},
  author = {Nathan Lambert and Jacob Morrison and Valentina Pyatkin and Shengyi Huang and Hamish Ivison and Faeze Brahman and Lester James V. Miranda and Alisa Liu and Nouha Dziri and Xinxi Lyu and Yuling Gu and Saumya Malik and Victoria Graf and Jena D. Hwang and Jiangjiang Yang and Ronan {Le Bras} and Oyvind Tafjord and Christopher Wilhelm and Luca Soldaini and Noah A. Smith and Yizhong Wang and Pradeep Dasigi and Hannaneh Hajishirzi},
  booktitle = {Second Conference on Language Modeling},
  year = {2025},
  url = {https://openreview.net/forum?id=i1uGbfHHpH}
}

@article{shao2024deepseekmath,
  title = {{DeepSeekMath}: Pushing the Limits of Mathematical Reasoning in Open Language Models},
  author = {Zhihong Shao and Peiyi Wang and Qihao Zhu and Runxin Xu and Junxiao Song and Xiao Bi and Haowei Zhang and Mingchuan Zhang and Y. K. Li and Y. Wu and Daya Guo},
  year = {2024},
  journal = {arXiv preprint arXiv:2402.03300},
  url = {https://arxiv.org/abs/2402.03300}
}

@inproceedings{hubotter2026sdpo,
  title = {Reinforcement Learning via Self-Distillation},
  author = {Jonas H{\"u}botter and Frederike L{\"u}beck and Lejs Behric and Anton Baumann and Marco Bagatella and Daniel Marta and Ido Hakimi and Idan Shenfeld and Thomas Kleine Buening and Carlos Guestrin and Andreas Krause},
  year = {2026},
  booktitle = {International Conference on Machine Learning},
  url = {https://openreview.net/forum?id=QkfkxyRizZ}
}

@article{ye2026oel,
  title = {Online Experiential Learning for Language Models},
  author = {Tianzhu Ye and Li Dong and Qingxiu Dong and Xun Wu and Shaohan Huang and Furu Wei},
  year = {2026},
  journal = {arXiv preprint arXiv:2603.16856},
  url = {https://arxiv.org/abs/2603.16856}
}

@inproceedings{yao2025taubench,
  title = {{$\tau$-bench}: A Benchmark for Tool-Agent-User Interaction in Real-World Domains},
  author = {Shunyu Yao and Noah Shinn and Pedram Razavi and Karthik Narasimhan},
  year = {2025},
  booktitle = {The Thirteenth International Conference on Learning Representations},
  url = {https://openreview.net/forum?id=roNSXZpUDN}
}

@article{yang2025qwen3,
  title = {{Qwen3} Technical Report},
  author = {An Yang and Anfeng Li and Baosong Yang and Beichen Zhang and Binyuan Hui and Bo Zheng and Bowen Yu and Chang Gao and Chengen Huang and Chenxu Lv and Chujie Zheng and Dayiheng Liu and Fan Zhou and Fei Huang and Feng Hu and Hao Ge and Haoran Wei and Huan Lin and Jialong Tang and Jian Yang and Jianhong Tu and Jianwei Zhang and Jianxin Yang and Jiaxi Yang and Jing Zhou and Jingren Zhou and Junyang Lin and Kai Dang and Keqin Bao and Kexin Yang and Le Yu and Lianghao Deng and Mei Li and Mingfeng Xue and Mingze Li and Pei Zhang and Peng Wang and Qin Zhu and Rui Men and Ruize Gao and Shixuan Liu and Shuang Luo and Tianhao Li and Tianyi Tang and Wenbiao Yin and Xingzhang Ren and Xinyu Wang and Xinyu Zhang and Xuancheng Ren and Yang Fan and Yang Su and Yichang Zhang and Yinger Zhang and Yu Wan and Yuqiong Liu and Zekun Wang and Zeyu Cui and Zhenru Zhang and Zhipeng Zhou and Zihan Qiu},
  year = {2025},
  journal = {arXiv preprint arXiv:2505.09388},
  url = {https://arxiv.org/abs/2505.09388}
}

@inproceedings{zhao2026opsd,
  title = {Self-Distilled Reasoner: On-Policy Self-Distillation for Large Language Models},
  author = {Siyan Zhao and Zhihui Xie and Mengchen Liu and Jing Huang and Guan Pang and Feiyu Chen and Aditya Grover},
  year = {2026},
  booktitle = {International Conference on Machine Learning},
  url = {https://openreview.net/forum?id=Jpxfof0EaS&noteId=qRewJdmNTV}
}

@inproceedings{shenfeld2026sdft,
  title = {Self-Distillation Enables Continual Learning},
  author = {Idan Shenfeld and Mehul Damani and Jonas H{\"u}botter and Pulkit Agrawal},
  year = {2026},
  booktitle = {International Conference on Machine Learning},
  url = {https://openreview.net/forum?id=qA6FgH0nnZ}
}

@article{yang2026ocsd,
  title = {Agentic Reinforcement Learning with Observation-Calibrated Self-Distillation},
  author = {Yi Yang and Cong Qin and Xiaodan Liu and Chishui Chen and Qing Dong and Yan Zhang and Cao Liu and Zhao Yang and Lu Pan and Jiaye Lin and Yi Feng},
  year = {2026},
  journal = {arXiv preprint arXiv:2608.04788},
  url = {https://arxiv.org/abs/2608.04788}
}

@article{liu2026hero,
  title = {{HERO: Hindsight-Enhanced Reflection from Environment Observations for Agentic Self-Distillation}},
  author = {Haoran Liu and Yuwei Zhang and Xiyao Li and Bohan Lyu and Jingbo Shang},
  year = {2026},
  journal = {arXiv preprint arXiv:2606.11559},
  url = {https://arxiv.org/abs/2606.11559}
}

@article{shrivastava2026echo,
  title = {ECHO: Terminal Agents Learn World Models for Free},
  author = {Vaishnavi Shrivastava and Piero Kauffmann and Ahmed Awadallah and Dimitris Papailiopoulos},
  year = {2026},
  journal = {arXiv preprint arXiv:2605.24517},
  url = {https://arxiv.org/abs/2605.24517}
}

@inproceedings{lu2026paw,
  title     = {Policy and World Modeling Co-Training for Language Agents},
  author    = {Ning Lu and Baijiong Lin and Shengcai Liu and Jiahao Wu and Haoze Lv and Yanbin Wei and Lingting Zhu and Shengju Qian and Xin Wang and Ying-Cong Chen and Qi Wang and Ke Tang},
  booktitle = {Proceedings of the 2026 Conference on Empirical Methods in Natural Language Processing},
  year      = {2026},
  url       = {https://arxiv.org/abs/2606.02388}
}

@article{cai2026aawm,
  title = {Beyond Next-Observation Prediction: Agent-Authored World Modeling for Sequential Decision Making},
  author = {Guangfeng Cai and Kaibing Yang and Shuo He and Yu Li and Shengtian Yang and Jiaqi Lv and Lei Feng},
  year = {2026},
  journal = {arXiv preprint arXiv:2606.25421},
  url = {https://arxiv.org/abs/2606.25421},
  doi = {10.48550/arXiv.2606.25421}
}

@inproceedings{agarwal2024onpolicy,
  title = {On-Policy Distillation of Language Models: Learning from Self-Generated Mistakes},
  author = {Rishabh Agarwal and Nino Vieillard and Yongchao Zhou and Piotr Stanczyk and Sabela Ramos and Matthieu Geist and Olivier Bachem},
  year = {2024},
  booktitle = {The Twelfth International Conference on Learning Representations},
  url = {https://proceedings.iclr.cc/paper_files/paper/2024/file/5be69a584901a26c521c2b51e40a4c20-Paper-Conference.pdf}
}

@inproceedings{tarvainen2017mean,
  title = {Mean Teachers Are Better Role Models: Weight-Averaged Consistency Targets Improve Semi-Supervised Deep Learning Results},
  author = {Antti Tarvainen and Harri Valpola},
  booktitle = {Advances in Neural Information Processing Systems},
  volume = {30},
  year = {2017},
  url = {https://proceedings.neurips.cc/paper/2017/hash/68053af2923e00204c3ca7c6a3150cf7-Abstract.html}
}

@inproceedings{gu2024minillm,
  title = {{MiniLLM}: Knowledge Distillation of Large Language Models},
  author = {Yuxian Gu and Li Dong and Furu Wei and Minlie Huang},
  booktitle = {The Twelfth International Conference on Learning Representations},
  year = {2024},
  url = {https://proceedings.iclr.cc/paper_files/paper/2024/hash/8ac015d409635f196f9e3e9dcfb9a94e-Abstract-Conference.html}
}

@inproceedings{hendrycks2021mmlu,
  title = {Measuring Massive Multitask Language Understanding},
  author = {Dan Hendrycks and Collin Burns and Steven Basart and Andy Zou and Mantas Mazeika and Dawn Song and Jacob Steinhardt},
  year = {2021},
  booktitle = {International Conference on Learning Representations},
  url = {https://openreview.net/forum?id=d7KBjmI3GmQ}
}

@inproceedings{wang2024mmlupro,
  title = {{MMLU-Pro}: A More Robust and Challenging Multi-Task Language Understanding Benchmark},
  author = {Yubo Wang and Xueguang Ma and Ge Zhang and Yuansheng Ni and Abhranil Chandra and Shiguang Guo and Weiming Ren and Aaran Arulraj and Xuan He and Ziyan Jiang and Tianle Li and Max Ku and Kai Wang and Alex Zhuang and Rongqi Fan and Xiang Yue and Wenhu Chen},
  year = {2024},
  booktitle = {Advances in Neural Information Processing Systems},
  volume = {37},
  url = {https://openreview.net/forum?id=y10DM6R2r3#discussion}
}

@article{zhou2023ifeval,
  title = {Instruction-Following Evaluation for Large Language Models},
  author = {Jeffrey Zhou and Tianjian Lu and Swaroop Mishra and Siddhartha Brahma and Sujoy Basu and Yi Luan and Denny Zhou and Le Hou},
  year = {2023},
  journal = {arXiv preprint arXiv:2311.07911},
  url = {https://arxiv.org/abs/2311.07911}
}

@inproceedings{chhikara2025mem0,
  title = {{Mem0}: Building Production-Ready {AI} Agents with Scalable Long-Term Memory},
  author = {Prateek Chhikara and Dev Khant and Saket Aryan and Taranjeet Singh and Deshraj Yadav},
  year = {2025},
  booktitle = {ECAI 2025},
  pages = {2993--3000},
  doi = {10.3233/FAIA251160},
  url = {https://doi.org/10.3233/FAIA251160}
}

@inproceedings{hu2022lora,
  title = {{LoRA}: Low-Rank Adaptation of Large Language Models},
  author = {Edward J. Hu and Yelong Shen and Phillip Wallis and Zeyuan Allen-Zhu and Yuanzhi Li and Shean Wang and Lu Wang and Weizhu Chen},
  booktitle = {International Conference on Learning Representations},
  year = {2022},
  url = {https://openreview.net/forum?id=nZeVKeeFYf9}
}

@article{tian2026autorubric,
  title   = {{Auto-Rubric} as Reward: From Implicit Preferences to Explicit Multimodal Generative Criteria},
  author  = {Tian, Juanxi and Liu, Fengyuan and Han, Jiaming and Jiang, Yilei and Wu, Yongliang and Liu, Yesheng and Li, Haodong and Xu, Furong and Li, Wanhua},
  journal = {arXiv preprint arXiv:2605.08354},
  year    = {2026},
  doi     = {10.48550/arXiv.2605.08354},
  url     = {https://arxiv.org/abs/2605.08354}
}

@article{liu2026dynacf,
  title   = {{DynaCF}: Mitigating Shortcut Learning in Reward Models via Dynamic Counterfactual Sensitivity},
  author  = {Liu, Fengyuan and Miao, Yongliang and He, Zirui and Liu, Yanguang and Sun, Fei and Du, Mengnan},
  journal = {arXiv preprint arXiv:2606.09043},
  year    = {2026},
  doi     = {10.48550/arXiv.2606.09043},
  url     = {https://arxiv.org/abs/2606.09043}
}
\bibliographystyle{iclr2027_conference}

\clearpage

\appendix
\providecommand{\selfpending}[1]{\textcolor{red}{\textbf{[AUTHOR TODO:} #1\textbf{]}}}
\providecommand{\selfmissing}{\textcolor{red}{Pending}}

\section{Datasets, Benchmarks, and Models}
\label{app:resources}

This section introduces the interactive environments,
general capability benchmarks, and models used to evaluate
SELF. Interactive environments assess learning from action
consequences, while general capability benchmarks measure
retention of knowledge, reasoning, and instruction following
after agent training.

\subsection{Datasets and Benchmarks}
\label{app:datasets}

\textbf{Interactive Benchmarks.}
We use $\tau$-bench Retail and AppWorld to evaluate learning
from agent--environment interactions. Learning uses the official
training splits, and task performance is measured on held-out
test splits. Training trajectories are collected by interacting
with the environments, rather than provided as expert
demonstrations. For SELF, the resulting observations serve as
feedback-prediction targets and as hindsight context for
self-distillation. Task-success annotations are reserved for
evaluation.

\begin{itemize}
    \item \textbf{$\tau$-bench Retail.}
    The retail domain of $\tau$-bench evaluates agents through
    multi-turn conversations with simulated users and calls
    to domain-specific tools~\citep{yao2025taubench}.
    Agents must interpret user requests, obtain relevant
    information, and perform actions consistent with retail
    policies. Tool responses reveal the consequences of actions
    and inform subsequent decisions, but successful execution
    of a tool call does not necessarily establish that the
    user's overall request has been fulfilled.
    We use Retail to assess learning from tool feedback in
    policy-constrained interactions.

    \item \textbf{AppWorld.}
    AppWorld evaluates agents that generate and execute code
    to complete tasks across multiple
    applications~\citep{trivedi2024appworld}.
    Tasks combine application APIs with program logic, and
    execution outputs guide subsequent actions. A single
    code-based action can compose multiple operations, creating
    dependencies both within actions and across interaction
    steps. Programmatic tests assess the resulting application
    states, including task completion and unintended changes.
    We use AppWorld to assess learning from code-execution
    feedback, complementing Retail's tool-based interactions.
\end{itemize}

\textbf{General Capability Benchmarks.}
MMLU, MMLU-Pro, and IFEval are used exclusively for evaluation.
We compare the initial Qwen3-4B-Instruct checkpoint, the trained
SELF student, and its EMA teacher to assess changes in general
capabilities after agent training. Each checkpoint receives
only the inputs provided by the corresponding benchmark.
These evaluations measure the teacher's general capabilities
separately from its use of privileged hindsight information
during interactive learning.

\begin{itemize}
    \item \textbf{MMLU.}
    MMLU evaluates knowledge and problem solving through
    multiple-choice questions across 57 subjects, including
    mathematics, history, computer science, and
    law~\citep{hendrycks2021mmlu}.
    Its broad subject coverage extends beyond the retail
    and application domains used for agent training.
    We use it to assess whether interactive adaptation
    preserves general knowledge and question-answering ability.

    \item \textbf{MMLU-Pro.}
    MMLU-Pro emphasizes reasoning-intensive questions and
    increases the number of answer options to as many as
    ten~\citep{wang2024mmlupro}.
    It complements MMLU with a more demanding assessment
    of reasoning and answer selection.
    We use it to examine whether these capabilities are
    retained after interactive training.

    \item \textbf{IFEval.}
    IFEval evaluates instruction following through
    programmatically verifiable requirements, such as
    constraints on response length and keyword
    usage~\citep{zhou2023ifeval}.
    It assesses whether generated responses satisfy explicit
    instructions without relying on a subjective model judge.
    We use it to measure retention of instruction-following
    behavior beyond tool use and code execution.
\end{itemize}

\subsection{Models}
\label{app:models}

We use Qwen3-4B-Instruct and Qwen3-8B as trainable
backbones to evaluate SELF across model scales,
and include
Qwen3-30B-A3B-Instruct as external comparisons.

\begin{itemize}
    \item \textbf{Qwen3-4B-Instruct.}
    Qwen3-4B-Instruct is an instruction-tuned language model
    in the Qwen3 family~\citep{yang2025qwen3}.
    We use it as a trainable backbone for the main
    interactive-task comparisons, coefficient ablations,
    mechanism analyses, and general capability evaluations.

    \item \textbf{Qwen3-8B.}
    Qwen3-8B is an open-weight language model in the
    Qwen3 family~\citep{yang2025qwen3}.
    We use its non-thinking configuration as a second
    trainable backbone to evaluate SELF on Retail and
    AppWorld at a larger model scale.

    \item \textbf{Qwen3-30B-A3B-Instruct.}
    Qwen3-30B-A3B-Instruct is an instruction-tuned
    mixture-of-experts model in the
    Qwen3 family~\citep{yang2025qwen3}.
    We report its performance on Retail and AppWorld
    to compare SELF with a larger model from the same family.
\end{itemize}
\section{Implementation Details}
\label{app:implementation}

This section specifies the training hyperparameters, dataset
partitions, and evaluation protocols used for SELF.
Benchmark and model descriptions are provided in
Appendix~\ref{app:resources}, and the training objectives
are derived in Appendix~\ref{app:objective}.

\subsection{Hyperparameters}
\label{app:hyperparameters}
\label{app:configuration}
\label{app:training}

\textbf{Optimization Settings.}
We train Qwen3-4B-Instruct and Qwen3-8B using LoRA~\citep{hu2022lora} with
AdamW~\citep{loshchilov2019decoupled}, a learning rate of $3\times10^{-4}$, and a training
batch size of 16. Qwen3-8B uses its non-thinking configuration.
The feedback-modeling and hindsight-distillation weights are
$\lambda_o=0.2$ and $\lambda_a=1.0$, respectively.
Both objectives update the same student adapters.
Table~\ref{tab:app_self_configuration} summarizes the default
configuration, including the backbone-specific LoRA settings.

\textbf{Interaction Collection and Optimization.}
Agents use a ReAct-style interface~\citep{yao2023react}, selecting actions from
the task instruction and accumulated interaction history.
At the start of each training round, the current student
collects trajectories containing transitions $(h_i,a_i,f_i)$,
where $h_i$ is the history preceding an action, $a_i$ is the
executed action, and $f_i$ is the resulting environmental
feedback. The collected trajectories remain fixed during
optimization within that round. The updated student then
collects the next round of interactions. Optimization samples a trajectory uniformly and then a turn
uniformly from that trajectory. Within each transition,
feedback-prediction and action-distillation losses are
normalized separately by their target lengths. This corresponds
to averaging over trajectories, turns, and target tokens,
rather than giving longer trajectories greater outer weight.
Recorded actions and feedback are treated as fixed data;
gradients do not propagate through sampling or environment
execution.

\begin{table}[t]
\caption{Default SELF training configuration. Backbone-specific
LoRA settings are listed separately; other settings are shared.}
\label{tab:app_self_configuration}
\begin{center}
\small
\setlength{\tabcolsep}{4pt}
\renewcommand{\arraystretch}{1.1}
\begin{tabular}{@{}lll@{}}
\toprule
\textbf{Hyperparameter} & \textbf{Qwen3-4B-Instruct} & \textbf{Qwen3-8B} \\
\midrule
Optimizer & \multicolumn{2}{l}{AdamW} \\
Learning rate & \multicolumn{2}{l}{$3\times10^{-4}$} \\
Training batch size & \multicolumn{2}{l}{16} \\
LoRA rank $r$ & 16 & 96 \\
LoRA alpha & 32 & 128 \\
LoRA dropout & 0.05 & 0.05 \\
Gradient checkpointing & \multicolumn{2}{l}{Enabled} \\
Feedback-modeling weight $\lambda_o$ & \multicolumn{2}{l}{0.2} \\
Hindsight-distillation weight $\lambda_a$ & \multicolumn{2}{l}{1.0} \\
Interaction interface & \multicolumn{2}{l}{ReAct-style} \\
Collection policy & \multicolumn{2}{l}{Current student at the start of each round} \\
Within-round data & \multicolumn{2}{l}{Fixed trajectory collection $\mathcal T_t$} \\
Action-prefix alignment & \multicolumn{2}{l}{Same recorded prefixes for student and teacher} \\
Teacher gradients & \multicolumn{2}{l}{Stopped during distillation} \\
Feedback-prediction loss & \multicolumn{2}{l}{Token-averaged negative log-likelihood} \\
Action-distillation loss & \multicolumn{2}{l}{Full-vocabulary reverse KL} \\
Loss averaging order & \multicolumn{2}{l}{Trajectories, turns, then target tokens} \\
Deployment & \multicolumn{2}{l}{Student action policy only} \\
\bottomrule
\end{tabular}
\end{center}
\end{table}

\textbf{Training Targets.}
Feedback prediction conditions on the history and executed
action, and uses the realized environmental response as its
target. During training, the response is scored through
teacher forcing with token-averaged negative log-likelihood.
For hindsight distillation, the student and teacher score the
same recorded action prefixes. The student receives the
available interaction history, while the teacher additionally
receives the realized feedback. Their token distributions
are aligned by position within the action, accounting for
the different conditioning-context lengths. Distillation uses reverse KL from the student distribution
to the detached teacher distribution over the full vocabulary~\citep{gu2024minillm}.
Feedback and action losses are evaluated only at their
respective target positions; conditioning and padding tokens
are excluded. At deployment, action generation uses only
the student policy, without an additional feedback-prediction
or teacher-scoring step.

\subsection{Dataset Setup}
\label{app:dataset-setup}

\textbf{Training and Test Partitions.}
We use the official training and test partitions of
$\tau$-bench Retail and AppWorld. Retail is evaluated on its official test split,
and AppWorld is evaluated on \texttt{test-normal}.
Training trajectories are collected through interactions
with training tasks, rather than provided as expert
solutions. Held-out test interactions do not contribute
to parameter updates.

\textbf{Supervision Available During Learning.}
SELF learns from environmental observations, including
ordinary execution outputs, error messages, and status
information. It does not add separate task-success labels
to either training objective. In the comparison settings,
Observation supplies environmental responses; Full additionally
supplies ground-truth information and explicit success/failure
feedback; Reward supplies verifiable task rewards.
OEL~\citep{ye2026oel} and SELF use Observation, SDPO~\cite{hubotter2026sdpo} and Mem0~\citep{chhikara2025mem0} are evaluated
under both Observation and Full, and GRPO~\citep{shao2024deepseekmath} uses Reward.
Task-success annotations remain available to the evaluator
for scoring held-out performance.

\textbf{Feedback-Prediction Evaluation Data.}
We construct a fixed transition collection for each interactive
benchmark using the initial Qwen3-4B-Instruct model. This model
runs over the benchmark's evaluation tasks three times,
producing recorded histories, actions, and actual environmental
responses. The resulting transitions are shared across the
feedback predictors being compared. These three collection
passes define the diagnostic evaluation data; they are distinct
from the four attempts per task used to calculate Retail
$\mathrm{pass}@4$.

\subsection{Evaluation Protocols}
\label{app:evaluation_protocols}

\textbf{Interactive Task Performance.}
The deployable student is evaluated on held-out tasks using
only the information available before each action.
Task completion is determined by the benchmark evaluator.
Retail reports success rate (SR), calculated as
$\mathrm{pass}@4$, and average interaction turns (Turns).
AppWorld reports its official Task Goal Completion (TGC)
and Scenario Goal Completion (SGC) metrics~\citep{trivedi2024appworld}.
SR, TGC, and SGC are expressed as percentages.

\textbf{Retail Success Rate and Interaction Length.}
For each of the $N$ test tasks, we conduct four independent
attempts. Let $s_{q,r}\in\{0,1\}$ indicate whether attempt
$r$ on task $q$ succeeds. A task contributes to
$\mathrm{pass}@4$ if at least one attempt succeeds:
\begin{equation}
    \mathrm{SR}
    =\mathrm{pass}@4
    =\frac{100}{N}\sum_{q=1}^{N}
      \mathbf 1\!\left[\sum_{r=1}^{4}s_{q,r}\geq1\right].
    \label{eq:app-self-pass4}
\end{equation}
Thus, each task receives equal weight, regardless of how many
of its attempts succeed. This measures success within four
attempts, rather than the proportion of tasks successful
in all four attempts.

Let $L_{q,r}$ be the interaction length of attempt $r$ on
task $q$. We average over all attempts, including failures:
\begin{equation}
    \mathrm{Turns}
    =\frac{1}{4N}\sum_{q=1}^{N}\sum_{r=1}^{4}L_{q,r}.
    \label{eq:app-self-turns}
\end{equation}
Turns is interpreted together with SR, since a short failed
interaction does not indicate efficient task completion.

\textbf{Feedback-Prediction Accuracy.}
Let $\mathcal E_b$ denote the fixed transition collection for
benchmark $b$ obtained from the three collection passes above.
For each recorded transition $(h_i,a_i,f_i)$, the evaluated
model generates a feedback prediction $\hat f_i$ conditioned
on $h_i$ and $a_i$. The realized response $f_i$ is withheld
from the predictor and used only as the reference for scoring.
This evaluation measures generated feedback, rather than
token likelihood under teacher forcing.

GPT-5.6-Luna, with reasoning effort set to \texttt{low},
compares $\hat f_i$ with $f_i$ and returns a binary judgment:
\begin{equation}
    c_i=J(\hat f_i,f_i)\in\{0,1\},
    \;
    \mathrm{Acc.}_b
    =\frac{100}{|\mathcal E_b|}
      \sum_{i\in\mathcal E_b}c_i.
    \label{eq:app-self-feedback-accuracy}
\end{equation}
Here, $c_i=1$ means that the judge considers the predicted
feedback consistent with the actual feedback, and $c_i=0$
means that it does not. The judge receives only the two
feedback texts, without the interaction history or action.
Consequently, Acc. measures agreement with observed feedback,
not whether the action itself is appropriate or completes
the task. All recorded steps receive equal weight, so longer
trajectories contribute more evaluation examples.
The same transition collection is used for every predictor
to separate prediction quality from differences in the
trajectories generated by their own policies.

\textbf{Privileged-Teacher Task Evaluation.}
The privileged teacher is evaluated on complete tasks,
using the same task-success criteria as the student.
At each decision step, we save the current environment state.
The student generates a trial action and executes it to obtain
environmental feedback. The teacher
then receives the student's pre-action context augmented
with that feedback. The student's trial action is not
included in the teacher prompt; the feedback is its only
additional information relative to the student's context.
The environment is then restored to the saved pre-action state,
and the teacher generates and executes its action from that state.
The resulting teacher-action transition is added to the ongoing
interaction history and determines the state for the next step.
The student trial is used only to obtain the additional feedback;
its state changes are rolled back. This process is repeated
until the task terminates, so task success is evaluated on
the trajectory executed by the teacher rather than by editing
a completed student trajectory.

Teacher task scores assess performance with privileged
feedback unavailable to the student at its original decision.
They are reported separately from deployable student scores
and do not represent ordinary inference without access to
additional environmental information.

\textbf{General Capability Retention.}
We evaluate the initial Qwen3-4B-Instruct model, the adapted
student, and the teacher checkpoint on MMLU, MMLU-Pro, and
IFEval. Each checkpoint receives the benchmark inputs without
privileged environmental feedback. These evaluations assess
knowledge, reasoning, and instruction following after agent
training and are excluded from the training objectives.
The teacher's general capability scores are therefore distinct
from its privileged interactive-task performance.

\section{Additional Results}
\label{app:additional-results}

\subsection{Effect of Cross-Rollout Feedback Modeling}
\label{app:cross-rollout-feedback}

\textbf{Experimental Design.}
We examine whether the benefit of environmental feedback modeling
depends on being coupled to the same interaction experience used for
hindsight self-distillation.
In standard SELF, the two objectives are computed from the same
recorded transitions: the realized feedback of a transition is used
both as privileged context for hindsight self-distillation and as the
prediction target for environmental feedback modeling.
We compare this setting with a cross-rollout variant, in which the
two objectives are trained on transitions from different rollouts.

Let
\[
\mathcal{B}
=
\{(h_i,a_i,f_i)\}_{i=1}^{B}
\]
denote a batch of transitions used for hindsight self-distillation.
Standard SELF also computes environmental feedback modeling on the
same batch,
\begin{equation}
\mathcal{L}_{\mathrm{SELF}}
=
\frac{\lambda_a}{B}
\sum_{i=1}^{B}
\ell_{\mathrm{sd}}(h_i,a_i,f_i)
+
\frac{\lambda_o}{B}
\sum_{i=1}^{B}
\ell_{\mathrm{env}}(h_i,a_i,f_i).
\label{eq:app-aligned-rollout}
\end{equation}

For the cross-rollout variant, we independently sample another batch
\[
\mathcal{B}'
=
\{(h'_i,a'_i,f'_i)\}_{i=1}^{B}
\]
from different collected rollouts and optimize
\begin{equation}
\mathcal{L}_{\mathrm{cross}}
=
\frac{\lambda_a}{B}
\sum_{i=1}^{B}
\ell_{\mathrm{sd}}(h_i,a_i,f_i)
+
\frac{\lambda_o}{B}
\sum_{i=1}^{B}
\ell_{\mathrm{env}}(h'_i,a'_i,f'_i).
\label{eq:app-cross-rollout}
\end{equation}
Importantly, each transition in $\mathcal{B}'$ preserves its original
history, action, and realized feedback.
Thus, the feedback-prediction task remains well defined:
$(h'_i,a'_i)\rightarrow f'_i$ is still an actual environment
transition.
The intervention only removes the requirement that feedback modeling
and hindsight self-distillation operate on the same interaction
experience.

This comparison allows us to distinguish the general benefit of
learning environmental feedback from the additional benefit of
coupling feedback modeling and hindsight supervision on the same
transitions.
All other training settings are kept fixed.
Each condition is evaluated in a single run.
We report Retail success rate (SR; $\mathrm{pass}@4$) and average
interaction turns, together with AppWorld Task Goal Completion (TGC)
and Scenario Goal Completion (SGC).

\begin{table}[t]
\caption{Effect of coupling environmental feedback modeling and
hindsight self-distillation on the same rollout with
Qwen3-4B-Instruct.
In the cross-rollout condition, feedback modeling uses intact
transitions from independently collected rollouts, while hindsight
self-distillation uses the original rollout.
Results are from one run per condition.
SR, TGC, and SGC are percentages, with higher values indicating
better performance.
Turns is interpreted alongside SR.
Bold values indicate the best completion scores.}
\label{tab:app-cross-rollout-feedback}
\begin{center}
\small
\setlength{\tabcolsep}{6pt}
\renewcommand{\arraystretch}{1.1}
\begin{tabular}{@{}lcccc@{}}
\toprule
\textbf{Condition}
& \multicolumn{2}{c}{\textbf{Retail}}
& \multicolumn{2}{c}{\textbf{AppWorld}} \\
\cmidrule(lr){2-3}
\cmidrule(lr){4-5}
& \textbf{SR $\uparrow$}
& \textbf{Turns $\downarrow$}
& \textbf{TGC $\uparrow$}
& \textbf{SGC $\uparrow$} \\
\midrule
Base & 27.1 & 10.7 & 17.00 & 5.56 \\
GRPO & 33.0 & 13.4 & 28.67 & 10.50 \\
SELF (cross-rollout) & 36.6 & 11.7 & 31.21 & 8.67 \\
\rowcolor{selfbg} SELF (same-rollout)
& \textbf{38.6} & \textbf{10.3}
& \textbf{32.16} & \textbf{12.44} \\
\bottomrule
\end{tabular}
\end{center}
\end{table}

\textbf{Results and Analysis.}
Table~\ref{tab:app-cross-rollout-feedback} shows that environmental
feedback modeling remains effective even when its training transitions
are decoupled from those used for hindsight self-distillation.
The cross-rollout variant achieves 36.6\% SR on Retail and
31.21\% TGC on AppWorld, substantially outperforming the unadapted
backbone and remaining competitive with the full SELF objective.
This result suggests that the benefit of feedback modeling does not
arise solely from using the same transition for both objectives:
learning to predict environmental responses from valid interaction
experience can improve the shared model even when that experience
comes from a different rollout.

Using the same rollout for both objectives provides an additional
benefit.
Compared with cross-rollout training, standard SELF improves Retail
SR from 36.6\% to 38.6\% while reducing average interaction length
from 11.7 to 10.3 turns.
On AppWorld, TGC increases from 31.21\% to 32.16\%, while SGC
increases from 8.67\% to 12.44\%.
These results indicate that environmental feedback modeling provides
a broadly useful learning signal on its own, while coupling it with
hindsight self-distillation on the same interaction experience yields
further gains.

\subsection{Training Cost}
\label{app:cost}

\textbf{Measurement Setup.}
We compare SELF with OEL~\citep{ye2026oel}, SDPO~\citep{hubotter2026sdpo} under the Observation setting,
and GRPO~\citep{shao2024deepseekmath} using four NVIDIA A100 GPUs 80\,GB.
For each method, we aggregate the training costs of its runs
in Table~1 across the reported backbones and interactive
benchmarks.
The accounting includes interaction collection and
optimization, but excludes final evaluation and additional
ablation experiments. Table~\ref{tab:app-self-training-cost} reports local GPU-hours
and wall time.
GPU-hours sum the device time used across runs, while wall
time sums the elapsed duration of the individual training
runs.

\begin{table}[t]
\caption{Training costs for the main experiments in Table~1
using four NVIDIA A100 80\,GB GPUs.
Values are aggregated across the reported training runs
and exclude final evaluation and additional ablations.
Lower values indicate lower cost.}
\label{tab:app-self-training-cost}
\begin{center}
\small
\setlength{\tabcolsep}{8pt}
\renewcommand{\arraystretch}{1.1}
\begin{tabular}{@{}lcc@{}}
\toprule
\textbf{Method}
& \textbf{GPU-hours}
& \shortstack{\textbf{Wall time}\textbf{(h)}} \\
\midrule
OEL & 20.3 & 6.12 \\
SDPO (Observation) & 11.8 & 3.67 \\
GRPO & 41.2 & 12.46 \\
\rowcolor{selfbg} SELF & 16.4 & 4.93 \\
\bottomrule
\end{tabular}
\end{center}
\end{table}

\textbf{Results and Analysis.}
SELF requires 16.4 GPU-hours and 4.93 hours of wall time.
Compared with GRPO, it reduces GPU usage by 60.2\% and
wall time by 60.4\%.
It also uses 19.2\% fewer GPU-hours and 19.4\% less wall
time than OEL.
Relative to SDPO (Observation), SELF incurs an additional
4.6 GPU-hours and 1.26 hours, corresponding to increases
of 39.0\% and 34.3\%, respectively. SELF therefore has a higher training cost than SDPO
(Observation), while remaining less expensive than OEL
and GRPO in these experiments.
Its feedback-prediction and hindsight-distillation objectives
reuse collected transitions without requiring additional
environment execution to construct their targets.
The additional computation is confined to training:
deployment uses only the student action policy.

\section{SELF Details}
\label{app:self_details}

This section presents the complete SELF training procedure,
then explains the construction of hindsight supervision and
the two learning objectives. We distinguish the data held fixed
within a collection round from the model quantities recomputed
at each optimization step, and specify how losses are aligned,
normalized, and differentiated.

\subsection{Training Workflow}
\label{app:self_workflow}

\textbf{Interaction Collection and Recorded Transitions.}
SELF alternates between collecting interactions and optimizing
the student. At the start of round $t$, the current student
interacts with the training environment through its deployable
action policy. A recorded transition $(h_i,a_i,f_i)$ contains
the task instruction and pre-action history $h_i$, the executed
action $a_i$, and the resulting environmental feedback $f_i$.
Earlier observations are included in $h_i$; the response to
$a_i$ becomes available only after that action is executed.

The resulting trajectory collection $\mathcal T_t$ remains
fixed throughout the round's optimization steps. SELF does not
resample actions or rerun the environment to construct targets
from these transitions. Instead, feedback prediction uses the
recorded response as a target, and hindsight distillation uses
that response as additional teacher context.
After optimization, the updated student collects the next round.

\textbf{Student and Teacher Roles.}
The student uses shared trainable parameters $\theta$ for action
generation and feedback prediction, with separate instructions
for the two uses. In the LoRA implementation, both objectives
update the same student adapters while the backbone weights
remain frozen. The teacher, denoted by parameters $\bar\theta$,
is initialized from the student and maintained through EMA
updates. Teacher outputs are detached during distillation.
Algorithm~\ref{alg:self} treats teacher maintenance as a separate
step rather than a gradient-based update through the teacher.

\textbf{Sampling and Optimization.}
Each minibatch contains 16 transitions. To give trajectories
equal outer weight, each transition is sampled by first selecting
a trajectory uniformly from $\mathcal T_t$ and then selecting a
turn uniformly from that trajectory. Both losses are computed
for each sampled transition and normalized by their respective
target lengths before averaging across the minibatch.
The resulting estimator matches the objective in
Equation~\ref{eq:app_finite_objective}.

Algorithm~\ref{alg:self} uses $K$ optimization steps per collection
round. AdamW updates only the student parameters; teacher outputs,
recorded prefixes, and environmental feedback are treated as
constants for that update. The teacher distributions are
recomputed using the current teacher checkpoint at subsequent
steps, even though the underlying transitions remain unchanged.
Default optimizer and LoRA settings are given in
appendix.

\begin{algorithm}[t]
\caption{SELF: training with environmental feedback and hindsight distillation}
\label{alg:self}
\small
\begin{algorithmic}[1]
\Require Initial student parameters $\theta_0$; training environment;
rounds $T$; updates per round $K$; batch size $B=16$
\Require Loss weights $\lambda_o=0.2$, $\lambda_a=1.0$;
learning rate $\eta=3\times10^{-4}$; student, teacher, and feedback prompts
\State Initialize student $\theta\gets\theta_0$ and teacher $\bar\theta\gets\theta_0$
\State Initialize the AdamW optimizer for the student trainable parameters
\For{$t=0,\ldots,T-1$}
    \State Collect $\mathcal T_t$ using the current student action policy
    \State Store each transition $(h_i,a_i,f_i)$; hold $\mathcal T_t$ fixed in this round
    \For{$s=1,\ldots,K$}
        \State Form $\mathcal B$ by sampling $B$ times: a uniform trajectory, then a uniform turn
        \ForAll{$(h_i,a_i,f_i)\in\mathcal B$}
            \State Compute student action distributions $p^{\mathrm{stu}}_{i,j}$ at recorded prefixes
            \State Compute detached teacher distributions $p^{\mathrm{tea}}_{i,j}$ using $h_i,f_i,a_{i,<j}$
            \State Align student and teacher outputs by action-token position $j$
            \State $\displaystyle \ell_{\mathrm{sd}}^{(i)}\gets
                \frac{1}{L_i}\sum_{j=1}^{L_i}
                D_{\mathrm{KL}}\!\left(p^{\mathrm{stu}}_{i,j}\middle\|
                \operatorname{sg}[p^{\mathrm{tea}}_{i,j}]\right)$
            \State Compute $p^{\mathrm{env}}_{i,k}$ using $h_i,a_i,f_{i,<k}$
            \State $\displaystyle \ell_{\mathrm{env}}^{(i)}\gets
                -\frac{1}{M_i}\sum_{k=1}^{M_i}
                \log p^{\mathrm{env}}_{i,k}(f_{i,k})$
        \EndFor
        \State $\displaystyle \widehat{\mathcal L}\gets
            \frac{1}{B}\sum_{i\in\mathcal B}
            \left(\lambda_o\ell_{\mathrm{env}}^{(i)}+
            \lambda_a\ell_{\mathrm{sd}}^{(i)}\right)$
        \State Clear student gradients and backpropagate $\widehat{\mathcal L}$
        \State Update student trainable parameters with AdamW at learning rate $\eta$
        \State Update the teacher using the maintained EMA rule, without backpropagation
    \EndFor
\EndFor
\State \Return Student action policy conditioned on the task and available history
\end{algorithmic}
\end{algorithm}

\subsection{Hindsight Self-Distillation}
\label{app:hindsight}

\textbf{Constructing Hindsight Supervision.}
Consider a collected transition $(h_i,a_i,f_i)$,
where $h_i$ contains the task instruction and the
interaction history before action $a_i$, and $f_i$
is the feedback returned after its execution.
The feedback records the realized consequence of
the action, but does not directly specify a corrected
action or a target policy.
SELF constructs such supervision through a hindsight
teacher that evaluates the recorded action with
access to both the original task context and its
observed consequence.

Let $a_i=(a_{i,1},\ldots,a_{i,L_i})$.
At action position $j$, the student and teacher
distributions are
\begin{align}
    p^{\mathrm{stu}}_{i,j}(v)
    &=
    \pi_\theta\!\left(
        v\mid h_i,a_{i,<j},
        \mathrm{prompt}_{\mathrm{stu}}
    \right),
    \label{eq:app_student_distribution}\\
    p^{\mathrm{tea}}_{i,j}(v)
    &=
    \pi_{\bar{\theta}}\!\left(
        v\mid h_i,f_i,a_{i,<j},
        \mathrm{prompt}_{\mathrm{tea}}
    \right),
    \label{eq:app_teacher_distribution}
\end{align}
for vocabulary token $v\in\mathcal{V}$.
The teacher parameters $\bar{\theta}$ are maintained
as an exponential moving average of the student.
The task instruction remains available through $h_i$,
so hindsight reassessment is conditioned on what the
agent was asked to accomplish, not merely on the
contents of the feedback.

The teacher and student evaluate the same recorded action
prefix $a_{i,<j}$. The complete action is not provided as
additional teacher context; at position $j$, only its prefix
is used to define the next-token prediction being compared.
The teacher additionally receives $f_i$, while the task and
pre-action history remain available through $h_i$.
This places both distributions at the same action-token
decision, with hindsight feedback as the teacher's privileged
information. The target is a learned distribution, not an
externally supplied corrected action.

Because feedback adds tokens to the teacher's input, the
student and teacher logits must be matched by position
within the recorded action, not by absolute position in
their full input sequences. Conditioning and padding positions
are excluded from the comparison. In the formulas below,
$L_i$ and $M_i$ denote the numbers of scored action and feedback
tokens, respectively, and are assumed to be positive.

\textbf{Action-Prefix Alignment and Distillation.}
SELF transfers hindsight supervision through the
length-normalized reverse KL
\begin{equation}
    \ell_{\mathrm{sd}}(h_i,a_i,f_i)
    =
    \frac{1}{L_i}
    \sum_{j=1}^{L_i}
    D_{\mathrm{KL}}\!\left(
        p^{\mathrm{stu}}_{i,j}
        \,\middle\|\,
        \operatorname{sg}
        [p^{\mathrm{tea}}_{i,j}]
    \right),
    \label{eq:app_distillation_loss}
\end{equation}
where $\operatorname{sg}$ denotes stop-gradient.
The comparison uses distributions over candidate
next tokens, rather than treating the recorded
token $a_{i,j}$ as a correct action label.
Consequently, the sampled action determines the
prefixes at which learning occurs without requiring
that the sampled action itself be successful.

For a fixed position, abbreviate
$p=p^{\mathrm{stu}}_{i,j}$ and
$q=\operatorname{sg}[p^{\mathrm{tea}}_{i,j}]$.
The local objective is
\begin{equation}
    D_{\mathrm{KL}}(p\|q)
    =
    \sum_{v\in\mathcal{V}}
    p(v)\bigl(\log p(v)-\log q(v)\bigr).
    \label{eq:app_reverse_kl}
\end{equation}
This objective allows the teacher to redistribute
support across the vocabulary.
Unlike hard-label imitation, it need not reinforce
the recorded next token; the update depends on the
teacher's distribution relative to the student's
current prediction.
The exact form of this update is derived below.

\textbf{Gradient Boundaries and Deployment.}
The observation $f_i$ enters the teacher context
during training, but not the deployable student's
context for action $a_i$.
Distillation therefore transfers information through
parameter updates rather than through access to
future observations at inference time.
Earlier observations remain available in $h_i$,
as in the original interaction process.

Within each gradient update, the recorded transition and
detached teacher prediction are fixed supervision.
Gradients do not pass through the teacher, action
sampling, or environment execution.
When trajectories are reused across optimization
steps, the recorded prefixes remain fixed even
as the student parameters change.
Equation~\ref{eq:app_distillation_loss} therefore
defines distillation at behavior-policy prefixes,
with the hindsight context supplied by each
recorded transition.

This training computation differs from the privileged-teacher
full-task evaluation in appendix.
Training scores recorded prefixes without executing teacher
actions. In that diagnostic evaluation, a student trial provides
feedback, the environment is restored to its pre-action state,
and the teacher generates and executes its own action.
Ordinary deployment uses neither this trial-and-restore procedure
nor future feedback.

\subsection{SELF Objective}
\label{app:objective}
We derive the feedback-prediction and hindsight-distillation
gradients, then express their contributions to the shared
student under the trajectory--turn--token weighting.

\textbf{Feedback Prediction as Forward KL.}
Hindsight supervision depends on the teacher's ability
to interpret the feedback it receives.
SELF complements this supervision with a predictive
objective that directly trains the shared model on
action--consequence relationships.

Write $f_i=(f_{i,1},\ldots,f_{i,M_i})$ and define
\begin{equation}
    p^{\mathrm{env}}_{i,k}(v)
    =
    p_{\mathrm{env},\theta}\!\left(
        v\mid h_i,a_i,f_{i,<k},
        \mathrm{prompt}_{\mathrm{env}}
    \right).
    \label{eq:app_feedback_token_distribution}
\end{equation}
These token distributions define the autoregressive
feedback model
\begin{equation}
    p_{\mathrm{env},\theta}
    (f_i\mid h_i,a_i,\mathrm{prompt}_{\mathrm{env}})
    =
    \prod_{k=1}^{M_i}
    p^{\mathrm{env}}_{i,k}(f_{i,k}).
    \label{eq:app_feedback_factorization}
\end{equation}
Let $\delta_{\mathrm{env},i,k}$ be a point mass on
the observed token $f_{i,k}$. This is an empirical token target
from a recorded response, rather than the full distribution
of possible environmental responses.
Its entropy is zero, giving
\begin{align}
    D_{\mathrm{KL}}\!\left(
        \delta_{\mathrm{env},i,k}
        \,\middle\|\,
        p^{\mathrm{env}}_{i,k}
    \right)
    &=
    \sum_{v\in\mathcal{V}}
    \delta_{\mathrm{env},i,k}(v)
    \log
    \frac{\delta_{\mathrm{env},i,k}(v)}
         {p^{\mathrm{env}}_{i,k}(v)}
    \notag\\
    &=
    -\log p^{\mathrm{env}}_{i,k}(f_{i,k}).
    \label{eq:app_forward_kl_token}
\end{align}
Here, zero-probability terms use the standard
convention $0\log 0=0$.
Averaging over the feedback tokens yields
\begin{align}
    \ell_{\mathrm{env}}(h_i,a_i,f_i)
    &=
    \frac{1}{M_i}
    \sum_{k=1}^{M_i}
    D_{\mathrm{KL}}\!\left(
        \delta_{\mathrm{env},i,k}
        \,\middle\|\,
        p^{\mathrm{env}}_{i,k}
    \right)
    \notag\\
    &=
    -\frac{1}{M_i}
    \sum_{k=1}^{M_i}
    \log p^{\mathrm{env}}_{i,k}(f_{i,k})
    \notag\\
    &=
    -\frac{1}{M_i}
    \log p_{\mathrm{env},\theta}
    (f_i\mid h_i,a_i,\mathrm{prompt}_{\mathrm{env}}).
    \label{eq:app_feedback_nll}
\end{align}
Thus, forward-KL feedback modeling is exactly
length-normalized maximum-likelihood learning from
the realized environment response.

For a feedback-prediction logit $u_{i,k,v}$,
softmax differentiation gives
\begin{equation}
    \frac{\partial\ell_{\mathrm{env}}}
         {\partial u_{i,k,v}}
    =
    \frac{1}{M_i}
    \left(
        p^{\mathrm{env}}_{i,k}(v)
        -
        \mathbf{1}\{v=f_{i,k}\}
    \right).
    \label{eq:app_feedback_gradient}
\end{equation}
For the response recorded in a transition, the target is available
regardless of whether the action ultimately contributes to task success.
For example, an execution error remains a valid
observation to predict.
The objective learns what follows an action,
rather than labeling the action as desirable.
This distinction explains why feedback modeling
complements, rather than replaces, policy learning.

\textbf{Deriving the Reverse-KL Gradient.}
At a fixed action position, write
$p=p^{\mathrm{stu}}_{i,j}$ and
$q=\operatorname{sg}[p^{\mathrm{tea}}_{i,j}]$.
Let $z_v$ denote the student logit, with
$p(v)=\exp(z_v)/\sum_u\exp(z_u)$.
Assuming positive probabilities, define
\begin{equation}
    A(v)=\log q(v)-\log p(v),
    \qquad
    \bar{A}
    =
    \sum_{u\in\mathcal{V}}p(u)A(u).
    \label{eq:app_relative_support}
\end{equation}
The quantity $A(v)$ measures relative teacher
support, and
$\bar{A}=-D_{\mathrm{KL}}(p\|q)$.
Since $q$ is detached,
\begin{equation}
    \frac{\partial D_{\mathrm{KL}}(p\|q)}
         {\partial p(u)}
    =
    \log p(u)-\log q(u)+1.
    \label{eq:app_kl_probability_gradient}
\end{equation}
Combining this expression with the softmax Jacobian,
\begin{equation}
    \frac{\partial p(u)}{\partial z_v}
    =
    p(u)\bigl(\mathbf{1}\{u=v\}-p(v)\bigr),
    \label{eq:app_softmax_jacobian}
\end{equation}
gives
\begin{align}
    \frac{\partial D_{\mathrm{KL}}(p\|q)}
         {\partial z_v}
    &=
    \sum_{u\in\mathcal{V}}
    p(u)\bigl(\mathbf{1}\{u=v\}-p(v)\bigr)
    \bigl(\log p(u)-\log q(u)+1\bigr)
    \notag\\
    &=
    p(v)\left[
        \log p(v)-\log q(v)
        -
        \sum_{u\in\mathcal{V}}
        p(u)\bigl(\log p(u)-\log q(u)\bigr)
    \right]
    \notag\\
    &=
    -p(v)\bigl(A(v)-\bar{A}\bigr).
    \label{eq:app_reverse_gradient}
\end{align}
The constant terms cancel because
$\sum_u p(u)=1$.
For the transition-level distillation loss,
the corresponding derivative is
\begin{equation}
    \frac{\partial\ell_{\mathrm{sd}}}
         {\partial z_{i,j,v}}
    =
    -\frac{1}{L_i}
    p^{\mathrm{stu}}_{i,j}(v)
    \left(
        A_{i,j}(v)-\bar{A}_{i,j}
    \right).
    \label{eq:app_distillation_gradient}
\end{equation}

If logits were optimized independently, gradient descent would
increase a token's logit when its relative teacher support
exceeds the student-weighted average, and decrease it when
that support is below average. In the actual model, these
logit derivatives propagate through shared trainable parameters,
so their effects across tokens and examples are coupled.
The centering arises from probability normalization,
not from a reward-based advantage estimator.
Although $A$ resembles an advantage notation,
it is computed entirely from the student and
teacher distributions.

This derivation explains how an observation can
produce dense policy supervision without specifying
a reference action.
The usefulness of the correction depends on whether the
feedback-conditioned teacher provides informative relative
support for candidate tokens.
The gradient expression characterizes how that
support is transferred; it does not establish that
every teacher reassessment improves task performance.

\textbf{Joint Objective and Effective Sample Weights.}
For a fixed collection $\mathcal{T}_t$, let $n_\tau$ be the
number of turns in trajectory $\tau$. Transition indices below
are local to each trajectory; the trajectory subscript on
$(h_i,a_i,f_i)$ is omitted for readability. SELF combines the
two transition-level objectives as follows:
\begin{equation}
    \mathcal{L}_{\mathrm{SELF},t}
    =
    \frac{1}{|\mathcal{T}_t|}
    \sum_{\tau\in\mathcal{T}_t}
    \frac{1}{n_\tau}
    \sum_{i=1}^{n_\tau}
    \left[
        \lambda_o\ell_{\mathrm{env}}(h_i,a_i,f_i)
        +
        \lambda_a\ell_{\mathrm{sd}}(h_i,a_i,f_i)
    \right].
    \label{eq:app_finite_objective}
\end{equation}
For nonempty trajectories and scored sequences,
the effective coefficients on individual feedback
and action token losses are, respectively,
\begin{equation}
    w^{\mathrm{env}}_{\tau,i,k}
    =
    \frac{\lambda_o}
         {|\mathcal{T}_t|\,n_\tau M_i},
    \qquad
    w^{\mathrm{sd}}_{\tau,i,j}
    =
    \frac{\lambda_a}
         {|\mathcal{T}_t|\,n_\tau L_i}.
    \label{eq:app_effective_weights}
\end{equation}
Each trajectory receives equal outer weight,
turns are averaged within their trajectory,
and each objective is normalized by its own
target length.
Long feedback sequences therefore do not receive
greater total weight merely because they contain
more tokens.
The coefficients $\lambda_o$ and $\lambda_a$
control the explicit weighting of the two
normalized objectives, although their gradient
magnitudes also depend on the model predictions.

This formulation corresponds to the
trajectory-then-turn sampling procedure in
Algorithm~\ref{alg:self}.
Uniformly sampling a flattened pool of transitions
would instead favor trajectories containing more
turns, unless equivalent weighting corrections
were applied.
Similarly, a single average over all tokens in
a batch generally differs from averaging the
length-normalized transition losses.

\textbf{Shared-Parameter Learning across Updates.}
Let
\begin{equation}
\begin{aligned}
    g_{\mathrm{env}}
    &=
    \frac{1}{|\mathcal{T}_t|}
    \sum_{\tau\in\mathcal{T}_t}
    \frac{1}{n_\tau}
    \sum_{i=1}^{n_\tau}
    \nabla_\theta\ell_{\mathrm{env}}^{(i)},
    \\
    g_{\mathrm{sd}}
    &=
    \frac{1}{|\mathcal{T}_t|}
    \sum_{\tau\in\mathcal{T}_t}
    \frac{1}{n_\tau}
    \sum_{i=1}^{n_\tau}
    \nabla_\theta\ell_{\mathrm{sd}}^{(i)}.
\end{aligned}
\label{eq:app_component_gradients}
\end{equation}
With the collected data and teacher outputs fixed
within the update, linearity gives
\begin{equation}
    \nabla_\theta\mathcal{L}_{\mathrm{SELF},t}
    =
    \lambda_o g_{\mathrm{env}}
    +
    \lambda_a g_{\mathrm{sd}}.
    \label{eq:app_joint_gradient}
\end{equation}
The two gradients act on the same student parameters,
but provide different supervision:
$g_{\mathrm{env}}$ learns the observed consequence
of the recorded action, whereas $g_{\mathrm{sd}}$
transfers the teacher's hindsight action distribution
into the deployable policy.

Their interaction also extends beyond the current
optimization step.
Feedback-modeling updates alter the student and,
through subsequent EMA updates~\citep{tarvainen2017mean}, the teacher used
for future hindsight supervision.
Policy distillation also changes these shared parameters
and can therefore affect feedback prediction, although the
direction of that effect is an empirical question.
It can additionally change the actions and
observations collected in later rounds.
Because the teacher is detached, this interaction
is not a same-step differentiable cycle.

This decomposition describes how the two objectives update
the shared model, while the coefficient studies provide
empirical evidence for their complementary benefits.
These benefits are not guaranteed by gradient addition alone;
the component gradients may reinforce or oppose one another
at different optimization steps.

\section{Illustrative Prompts and Examples}

\begin{historybox}

\trajlabel{$q$} \textbf{Task instruction}

Fix the Python program \texttt{analyze\_sales.py}.
The script should read \texttt{data/orders.csv}, consider only
orders whose status is \texttt{completed}, compute the average
revenue of those orders, and print the result with exactly two
decimal places.

You may inspect files, execute Python or shell commands, and edit
the implementation. Do not modify the tests.

\medskip

\trajlabel{$o_0$} \textbf{Initial environment observation}

Working directory: \texttt{/workspace/sales\_analysis}

Repository contents:

\texttt{
analyze\_sales.py \\
data/orders.csv \\
tests/test\_analyze\_sales.py \\
README.md
}

\medskip

\trajlabel{$a_1$} \textbf{Agent action}

\texttt{sed -n '1,220p' analyze\_sales.py}

\medskip

\trajlabel{$f_1$} \textbf{Environmental feedback}

\texttt{
import csv \\[1mm]
\\
def main(): \\
\hspace*{4mm}total = 0.0 \\
\hspace*{4mm}count = 0 \\
\hspace*{4mm}with open("data/orders.csv") as f: \\
\hspace*{8mm}reader = csv.DictReader(f) \\
\hspace*{8mm}for row in reader: \\
\hspace*{12mm}total += float(row["revenue"]) \\
\hspace*{12mm}count += 1 \\
\hspace*{4mm}print(f"Average revenue: \{total/count:.1f\}") \\
\\
if \_\_name\_\_ == "\_\_main\_\_": \\
\hspace*{4mm}main()
}

\medskip

\trajlabel{$a_2$} \textbf{Agent action}

\texttt{head -5 data/orders.csv \&\& pytest -q}

\medskip

\trajlabel{$f_2$} \textbf{Environmental feedback}

\texttt{
order\_id,status,total \\
1001,completed,32.50 \\
1002,cancelled,80.00 \\
1003,completed,62.00 \\
1004,pending,44.00
}

\smallskip

\texttt{
F \\
============================= FAILURES ============================= \\
KeyError: 'revenue' \\
analyze\_sales.py:10
}

\medskip

\trajlabel{$a_3$} \textbf{Agent action}

\texttt{sed -n '1,220p' tests/test\_analyze\_sales.py}

\medskip

\trajlabel{$f_3$} \textbf{Environmental feedback}

\texttt{
def test\_completed\_orders\_only(capsys): \\
\hspace*{4mm}main() \\
\hspace*{4mm}out = capsys.readouterr().out.strip() \\
\hspace*{4mm}assert out == "Average completed-order revenue: 47.25"
}

\medskip

\trajlabel{$a_4$} \textbf{Agent action}

Inspect all rows to verify which records should contribute:

\texttt{cat data/orders.csv}

\medskip

% \trajlabel{$f_4$} \textbf{Environmental feedback}

% \texttt{
% order\_id,status,total \\
% 1001,completed,32.50 \\
% 1002,cancelled,80.00 \\
% 1003,completed,62.00 \\
% 1004,pending,44.00 \\
% 1005,refunded,15.00
% }

% \medskip

% \metatext{
% ... several local reasoning and inspection steps omitted ...
% }

% \medskip

% \trajlabel{$a_5$} \textbf{Agent action}

% Edit \texttt{analyze\_sales.py} to filter completed orders
% and use the correct column:

% \texttt{
% python - <<'PY' \\
% from pathlib import Path \\
% p = Path("analyze\_sales.py") \\
% s = p.read\_text() \\
% s = s.replace('total += float(row["revenue"])', \\
% \hspace*{15mm}'if row["status"] != "completed": continue\textbackslash n'
% \hspace*{15mm}'        total += float(row["total"])') \\
% s = s.replace('count += 1', \\
% \hspace*{15mm}'        count += 1') \\
% s = s.replace('Average revenue:', \\
% \hspace*{15mm}'Average completed-order revenue:') \\
% p.write\_text(s) \\
% PY
% }

% \medskip

% \trajlabel{$f_5$} \textbf{Environmental feedback}

% Command completed successfully. File updated.

% \medskip

% \trajlabel{$a_6$} \textbf{Agent action}

% \texttt{pytest -q}

% \medskip

\ldots

\trajlabel{$f_6$} \textbf{Environmental feedback}

\texttt{
F \\
============================= FAILURES ============================= \\
Expected: Average completed-order revenue: 47.25 \\
Received: Average completed-order revenue: 47.2 \\
1 failed in 0.08s
}

\medskip

\trajlabel{$a_7$} \textbf{Agent action}

\texttt{
python - <<'PY' \\
from pathlib import Path \\
p = Path("analyze\_sales.py") \\
s = p.read\_text().replace(':.1f', ':.2f') \\
p.write\_text(s) \\
PY
}

\medskip

\trajlabel{$f_7$} \textbf{Environmental feedback}

Command completed successfully.

\medskip

\trajlabel{$a_8$} \textbf{Agent action}

\texttt{pytest -q}

\medskip

\trajlabel{$f_8$} \textbf{Environmental feedback}

\texttt{
. \\
1 passed in 0.07s
}

\medskip

\trajlabel{$a_9$} \textbf{Agent action}

\texttt{python analyze\_sales.py}

\medskip

\trajlabel{$f_9$} \textbf{Environmental feedback}

\texttt{
Average completed-order revenue: 47.25
}

\medskip

\trajlabel{$a_{10}$} \textbf{Agent action}

\texttt{git diff -- analyze\_sales.py}

\medskip

\trajlabel{$f_{10}$} \textbf{Environmental feedback}

\texttt{
- total += float(row["revenue"]) \\
- count += 1 \\
+ if row["status"] != "completed": \\
+\hspace*{4mm}continue \\
+ total += float(row["total"]) \\
+ count += 1 \\
- print(f"Average revenue: \{total/count:.1f\}") \\
+ print(f"Average completed-order revenue: \{total/count:.2f\}")
}

\medskip

\trajlabel{$a_n$} \textbf{Final agent response}

Implemented the fix in \texttt{analyze\_sales.py}.
The program now filters for completed orders, uses the
\texttt{total} field, and prints the average with two decimal
places. The test suite passes successfully.

\end{historybox}

\clearpage

\vspace{3mm}
% Example 1
% =========================================================
\begin{predictionbox}{purplebg}

\field{Example 1 --- Python execution \hfill \textcolor{blue}{Correct} prediction}

\medskip

\field{Task $q$:}
Debug a Python script that parses a JSONL dataset.

\smallskip

\field{History $h_i$:}
\ldots\ The agent inspected the schema and found that each record
contains \texttt{id}, \texttt{text}, and \texttt{label}.

\smallskip

\field{Current action $a_i$:}
\texttt{python validate\_dataset.py data/train.jsonl}

\smallskip

\field{Realized feedback $f_i$:}
\texttt{KeyError: 'user\_id' at validate\_dataset.py:27}

\smallskip

\field{Predicted feedback $\hat f_i$:}
\texttt{KeyError: 'user\_id' at validate\_dataset.py:27}

\smallskip

\status{Result:} The prediction correctly identifies both the
exception type and the missing field.

\end{predictionbox}

\vspace{4mm}

\begin{predictionbox}{greenbg}

\field{Example 2 --- Retail tool interaction \hfill \textcolor{red}{Incorrect} prediction}

\medskip

\field{Task $q$:}
Determine whether the user's wireless-headphone order
is eligible for return.

\smallskip

\field{History $h_i$:}
\ldots\ The user account has been verified and Order 1042
was identified as the relevant purchase.

\smallskip

\field{Current action $a_i$:}
\texttt{get\_order\_details(order\_id="1042")}

\smallskip

\field{Realized feedback $f_i$:}
Order 1042 was delivered 12 days ago and remains within
the return window.

\smallskip

\field{Predicted feedback $\hat f_i$:}
Order 1042 was cancelled before shipment and is eligible
for an automatic refund.

\smallskip

\status{Result:} The model predicts the wrong order state
and therefore misinterprets the consequence of the action.

\end{predictionbox}

\vspace{4mm}
% ========================================================
% Example 3
% =========================================================
\begin{predictionbox}{purplebg}

\field{Example 3 --- SQL database interaction \hfill \textcolor{red}{Incorrect} prediction}

\medskip

\field{Task $q$:}
Count the number of pending orders in the database.

\smallskip

\field{History $h_i$:}
\ldots\ The agent connected successfully to
\texttt{orders.db} and verified the table schema.

\smallskip

\field{Current action $a_i$:}
\texttt{SELECT COUNT(*) FROM orders WHERE status='pending';}

\smallskip

\field{Realized feedback $f_i$:}
\texttt{17}

\smallskip

\field{Predicted feedback $\hat f_i$:}
\texttt{0}

\smallskip

\status{Result:} The predicted database response is incorrect
despite the action itself being syntactically valid.

\end{predictionbox}

\vspace{4mm}
% =========================================================
% Example 4
% =========================================================
\begin{predictionbox}{greenbg}

\field{Example 4 --- Filesystem interaction \hfill \textcolor{blue}{Correct} prediction}

\medskip

\field{Task $q$:}
Locate the configuration used by the training script.

\smallskip

\field{History $h_i$:}
\ldots\ The repository contains \texttt{train.py},
\texttt{configs/}, and several experiment directories.

\smallskip

\field{Current action $a_i$:}
\texttt{cat configs/base.yaml}

\smallskip

\field{Realized feedback $f_i$:}
\texttt{model: qwen3-4b} \\
\texttt{batch\_size: 16} \\
\texttt{learning\_rate: 1e-6}

\smallskip

\field{Predicted feedback $\hat f_i$:}
\texttt{model: qwen3-4b} \\
\texttt{batch\_size: 16} \\
\texttt{learning\_rate: 1e-6}

\smallskip

\status{Result:} The predicted environmental feedback matches
the observed file contents.

\end{predictionbox}

\subsection{Prompts}

\begin{figure}[h]
    \centering
    \includegraphics[width=\linewidth]{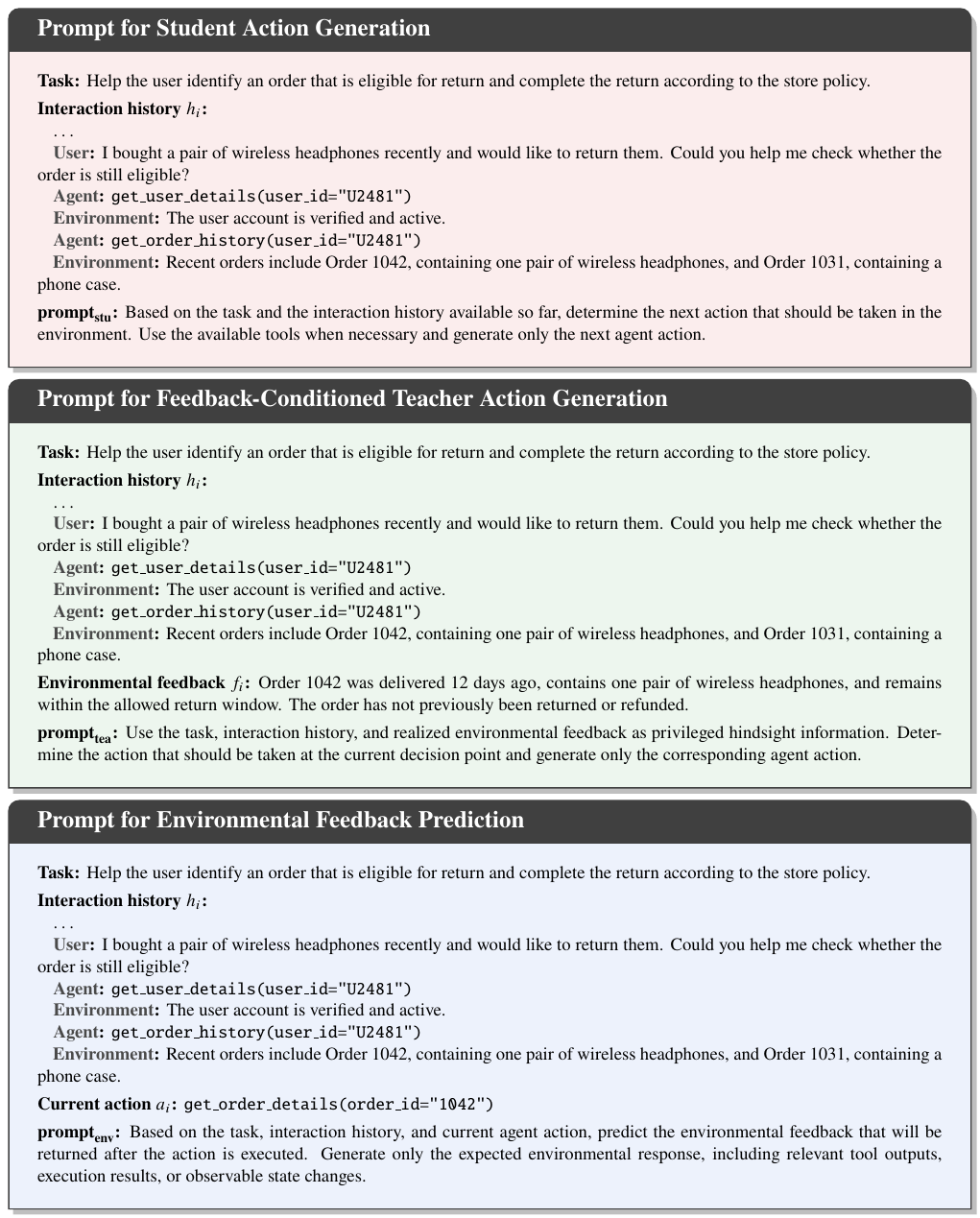}
    \caption{
    Illustrative prompting interfaces used in SELF.
    The student prompt $\mathrm{prompt}_{\mathrm{stu}}$ generates
    the next action $a_i$ from the interaction history $h_i$;
    the teacher prompt $\mathrm{prompt}_{\mathrm{tea}}$ additionally
    conditions on the realized environmental feedback $f_i$ as
    privileged hindsight information; and the environmental-feedback
    modeling prompt $\mathrm{prompt}_{\mathrm{env}}$ predicts $f_i$
    from $h_i$ and the current action $a_i$.
    }
    \label{fig:self_prompts}
\end{figure}

\end{document}